\documentclass{article}

\usepackage[preprint]{neurips_2025}

\usepackage[utf8]{inputenc} 
\usepackage[T1]{fontenc}    
\usepackage{hyperref}       
\usepackage{url}            
\usepackage{booktabs}       
\usepackage{amsfonts}       
\usepackage{nicefrac}       
\usepackage{microtype}      
\usepackage{xcolor}         

\usepackage{graphicx}
\usepackage{amsmath}
\usepackage{amssymb}
\usepackage{subfig}
\usepackage{multicol}
\usepackage{multirow}

\expandafter\def\expandafter\normalsize\expandafter{%
    \normalsize%
    \setlength\abovedisplayskip{3pt}%
    \setlength\belowdisplayskip{3pt}%
}

\title{FastFace: Tuning Identity Preservation in Distilled Diffusion via Guidance and Attention}

\author{%
  Sergey Karpukhin \\
  Skoltech\thanks{Skolkovo Institute of Science and Technology}, AIRI \\
  Moscow, Russia \\
  \texttt{Karpukhin@airi.net} \\
\And
  Vadim Titov \\
  AIRI \\
  Moscow, Russia \\
  \texttt{titow2408@gmail.com} \\
\AND
  Andrey Kuznetsov \\
  AIRI, Sber, Innopolis \\
  Moscow, Russia \\
  \texttt{Kuznetsov@airi.net} \\
\And
  Aibek Alanov \\
  HSE University, AIRI \\
  Moscow, Russia \\
  \texttt{alanov.aibek@gmail.com} \\
}

\begin{document}

\maketitle

\begin{figure}[!htb]
  \centering
  \includegraphics[width=1.0\textwidth]{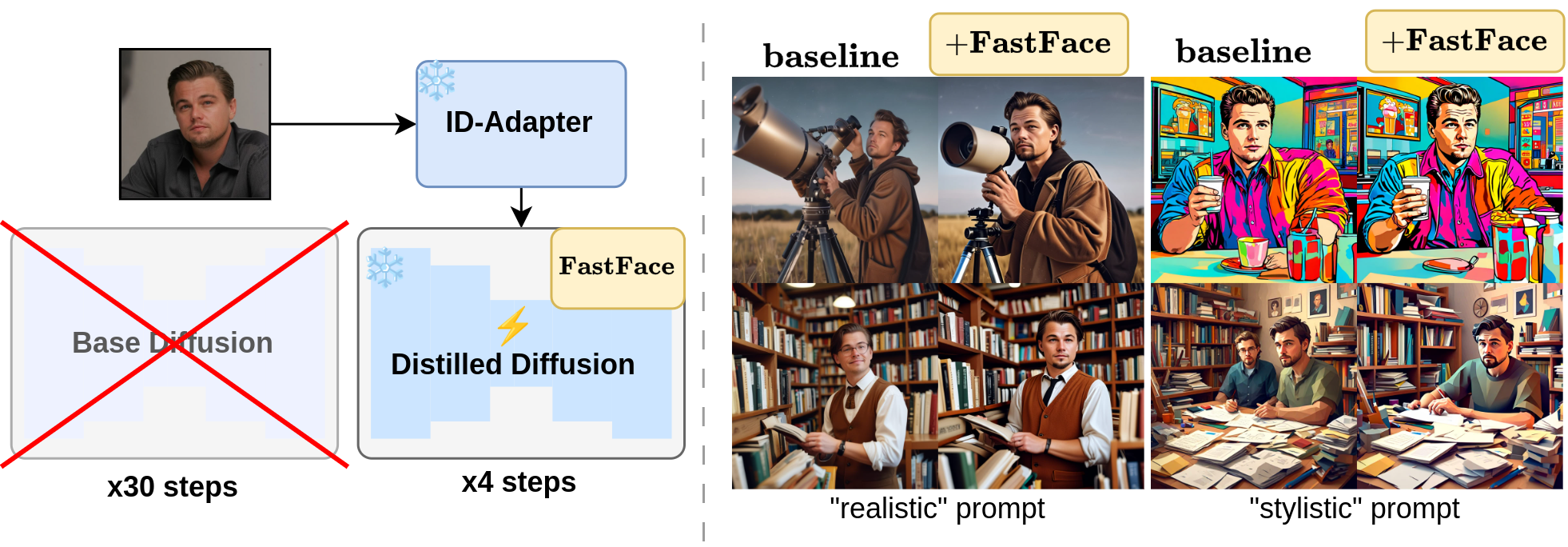}
  \caption{FastFace method framework: on the left - high-level idea of pipeline, enabling few-sep id-preserving generation, on the right - effect of FastFace components on realistic and stylistic generations} 
\label{fig: fastface_framework}
\end{figure}
\raggedbottom

\begin{abstract}
    In latest years plethora of identity-preserving adapters for a personalized generation with diffusion models have been released. Their main disadvantage is that they are dominantly trained jointly with base diffusion models, which suffer from slow multi-step inference. This work aims to tackle the challenge of training-free adaptation of pretrained ID-adapters to diffusion models accelerated via distillation - through careful re-design of classifier-free guidance for few-step stylistic generation and attention manipulation mechanisms in decoupled blocks to improve identity similarity and fidelity, we propose universal FastFace framework. Additionally, we develop a disentangled public evaluation protocol for id-preserving adapters.
\end{abstract}

\section{Introduction}

Diffusion models have emerged as a dominant paradigm in generative modeling, achieving state-of-the-art performance in high-fidelity image synthesis, with plethora of models coming out in recent years (\cite{ho2020ddpm}, \cite{dhariwal2021diffusion}, \cite{rombach2022high}, \cite{podell2023sdxl}, \cite{esser2024scaling}, \cite{flux2024}). Their iterative denoising process enables fine-grained control over generation but comes at the cost of slow inference bottleneck. This problem has been studied in the context of diffusion acceleration via distillation, with a lot of approaches and versions releasing in past several years, such as LCM, Turbo, Lightning, Hyper, and others (\cite{luo2023latent}, \cite{sauer2024fast}, \cite{lin2024sdxl}, \cite{ren2024hyper}, \cite{sauer2024adversarial}); common results of these distillation are 1) architecture of diffusion model remains the same 2) inference becomes significantly more efficient in terms of number of steps. In parallel, diffusion models have been adapted for controllable generation with image condition, in particular for the task of id-preserving generation, where conditional image contains face of a person, and diffusion can generate images with novel identities without further finetuning (\cite{ye2023ip}, \cite{li2024photomaker}, \cite{wang2024instantid}, \cite{guo2024pulid}, \cite{jiang2025infiniteyou}). These methods are usually originally trained and integrated with full-step diffusion models (with the exception of PuLID), however, their interaction with distilled versions introduces new challenges and opportunities. While most of these methods can be applied out of the box to distilled versions, their application may introduce instabilities or sub-optimal performance given basic tuning options.

Integrating personalized adapters with distilled diffusion models is highly relevant for advancing the practical application of id-preserving generation to real-time systems, aiming for efficiency and user responsiveness. Similar problem setup has been explored in literature in the setting of ControlNet (\cite{zhang2023adding}, \cite{xiao2023ccm}, \cite{parmar2024one}), proposing specific per-model finetuning strategy to adapt prior of ControlNet towards new trajectories of distilled models. Despite these studies exploring options for finetuning, they don't propose a universal approach to any model, i.e. for a new model, a completely new algorithmic design is required. While there are adaption studies with general setting of finetuning to any diffusion models, they do not consider distilled setup \cite{lin2024ctrl}. When new id-preserving methods are released for base models, there is a substantial need for methods that would generalize across distilled versions of the model out of the box, preferably without additional training by the end-user. 

Instead of going for separate, per-model finetuning approach, we aim to develop universal, training-free mechanisms that can be used in plug-and-play manner to improve quality of id-preserving generation with any distilled diffusion model. We start by separating context of application of ID-adapters - realistic and stylistic generations, and develop mechanisms in context of these setups. For stylistic generation we propose and tune decoupled classifier-free guidance, where conditional noise prediction is splitted into two parallel terms, and apply scheduling and rescaling targeted for few-step inference of distilled diffusion model. Secondly, independently we propose general approach of attention manipulation, where attention in decoupled blocks is transformed to be more focused on facial regions during generation, enhancing identity-preserving properties, and design two different transforms with trade-offs between each other - scale-power and scheduled-softmask. First transform is tuned for local identity preservation enhancement, while second is designed to bias generation towards more stable, portrait-like images with larger faces. Joint application of these mechanisms is denoted as FastFace framework and is summarized in Figure \ref{fig::general_method_scheme}. Additionally, we construct our case-specific dataset of identity images and prompts, allowing tuning each setting separately. In the end, we evaluate both proposed methods across joint setting of stylistic and realistic generation as a general framework and demonstrate its superior qualities in terms of metrics scaling.  

Overall, this work proposes the following contributions: 

\noindent\underline{Decoupled Classifier-Free Guidance Mechanism}: a training-free guidance strategy is introduced, decomposing classifier-free guidance into semantically interpretable components, and tuned to work best in the few-step sampling regime of distilled models. 

\noindent\underline{Attention Manipulation for Identity Enhancement}: inference-time method is developed to manipulate attention maps in decoupled attention blocks. By carefully manipulating values in attention maps it is reinforced over facial regions, substantially improving identity similarity without additional training and demonstrating robust performance.
    
\noindent\underline{Evaluation Protocol for Identity-Preserving Generation}: a systematic open evaluation protocol is proposed, which disentangles problems of "stylistic" and "realistic" generation cases and allows to purposefully tune id-preserving adapters towards one or another.

\section{Related work}

\paragraph{ID-preserving generation methods}

Identity-preserving generation, as we describe it, is a problem of preserving identity similarity in generation output given an image with the face. A lot of methods came out around this problem, including IpAdapter-FaceID \cite{ye2023ip}, Photomaker \cite{li2024photomaker}, PuLID \cite{guo2024pulid}, InstantID \cite{wang2024instantid}. They differ in their overall approaches and flexibility, with later methods building on top of FaceID, however, id-adapters trained for new diffusion models frequently rely on conventional FaceID approach and codebase (\cite{kolors}). Another group of methods such as DreamBooth \cite{ruiz2023dreambooth} and similar are also applicable to this problem, however, they are heavily limited due to need for finetuning for each new identity.

\paragraph{Diffusion distillation}

Diffusion distillation is an approach to accelerate trained diffusion models by training them to sample in few steps while still trying to model original $p_{data}(x)$ as close as possible (\cite{salimans2022progressive}, \cite{song2023consistency}, \cite{yin2024one}). State of the art approaches such as LCM \cite{luo2023latent} and Hyper \cite{ren2024hyper} remain common for new model releases (\cite{ke2023repurposing}, \cite{chen2024pixart})
), but new distillation techniques are actively being developed. In practice, these distilled versions may differ in their inference qualities and sampling procedures, generally applicable in range of 1-8 sampling steps. Application of these distilled models to image conditioned generation and in particular id-preserving generation is at the heart of this work. 

\paragraph{Adaptation to new diffusion models}

Cheap adaptation of pretrained modules for diffusion models to new checkpoints has been explored in \cite{lin2024ctrl} authors train an adapter module that acts as a latent projection between the inner-layer connection of the original ControlNet and new diffusion model and they achieve fast generalization. In other work \cite{xu2024ctrlora} authors consider a case of efficient adaptation of ControlNet to new conditional domains. In the context of distilled diffusion models similar problems have also been explored with ControlNet: (\cite{xiao2023ccm}, \cite{parmar2024one}), where in both works specific finetuning approaches are proposed either to match distillation objective or enforce cycle-consistency. Limitations of available solutions are either designing finetuning approach per checkpoint or tolerating baseline quality. We show that it is possible to universally boost quality of such adaptation without any additional training.

\section{Preliminary}

\paragraph{General problem formulation} Diffusion models are trained on the task of denoising image data. Common optimization problem is given in Eq. \ref{eq::diffusion_target}, which corresponds to variance preserving forward noising of data and prediction of added noise to an image. Additionally, during training a caption condition $c_{text}$ is added, which later allows to condition on text to generate image.

\begin{equation}\label{eq::diffusion_target}
    \theta^\star = \arg\min_\theta\mathcal{L} = \mathbb{E}_{x,t,\epsilon}[\| \epsilon - \epsilon_\theta(\sqrt a_t x + \sqrt{1- a_t}\epsilon, t, c_{text}) \|^2]
\end{equation}

For clarity of formulation we can denote $D_\theta(x_t, t, c_{text})$ a denoiser network, a function which predicts less noised $x_{t'}$, where $t' < t$ are timesteps defined by sampling schedule, denoted as $\{t_{i}\}^N_{i=0}$. Then, an adapter module $\phi$ (which in particular case is id-preserving adapter) is trained jointly with original model, which we denote through union $\theta \cup \phi$, optimization target is given in Eq. \ref{eq::diffusion_adapter_target}. Result of such training is a denoiser that accepts image condition $c_{img}$ - $D_{\theta \cup \phi}(x_t, t, c_{text}, c_{img})$.

\begin{equation}\label{eq::diffusion_adapter_target}
   \phi = \arg\min_\phi\mathcal{L} = \mathbb{E}_{x,t,\epsilon}[\| \epsilon - \epsilon_{stopgrad(\theta) \cup \phi}(\sqrt a_t c_{img} + \sqrt{1- a_t}\epsilon, t, c_{text}, c_{img}) \|^2]
\end{equation}

Adapter module is trained jointly with frozen parameters of the original model $\theta$. It is important to note that usually same condition image $c_{img}$ is used as denoising target and condition input, however last is processed by separate encoder. In context of id-preserving adapters, we denote $c_{img}$ as $c_{id}$, emphasizing that this is an image with persons face in it. This work studies application of pretrained id-preserving adapter to distilled versions of the original model $\theta$, denoted as $\tilde{\theta}$, trained to reduce the number of steps needed for sampling. Distilled model inference is defined by new timestep schedule $\{\tilde t_{i}\}^M_{i=0}$, where $M \ll N$, while model architecture is kept without changes.  Distillation procedure is abstracted, leaving us with $D_{\tilde{\theta}}(x_t, t, c_{text})$ and $D_{\tilde{\theta} \cup \phi}(x_T, T, c_{text}, c_{id})$ for text and id-conditioned denoisers.

\section{Method}

Goal of this work is to introduce plug-and-play, training-free mechanisms that allow to effectively tune any id-preserving adapter with distilled diffusion models and study their scaling behaviors w.r.t. conventional hyper-parameter \texttt{ip\_adapter\_scale} or $\lambda$, which impacts conditioning strength through interaction of attention and decoupled attention in Eq. \ref{eq::adapterscale}. where and $K'$ and $V'$ are keys and values in decoupled blocks (for details see \cite{ye2023ip}). 
\begin{equation}\label{eq::adapterscale}
  z_{new} = Attention(z; Q, K, V) + \lambda \cdot Attention(z; Q, K', V') 
\end{equation}

In context this work under specified problem we limit scope of experimental setup to models listed below. Idea is to demonstrate universality of proposed methods to different distilled checkpoints of base multi-step diffusion model:

\begin{itemize}
    \item $D_\theta$ - StableDiffusionXL (SDXL)
    \item $D_{\theta \cup \phi}$ - SDXL + IpAdapter FaceID-Plus-v2
    \item $D_{\tilde{\theta}}$ - SDXL-Turbo, SDXL-LCM, SDXL-Lightning, SDXL-Hyper, all checkpoints are used only in 4-step sampling regime
\end{itemize}

\begin{figure}[!htb]
  \centering
  \includegraphics[width=1.0\textwidth]{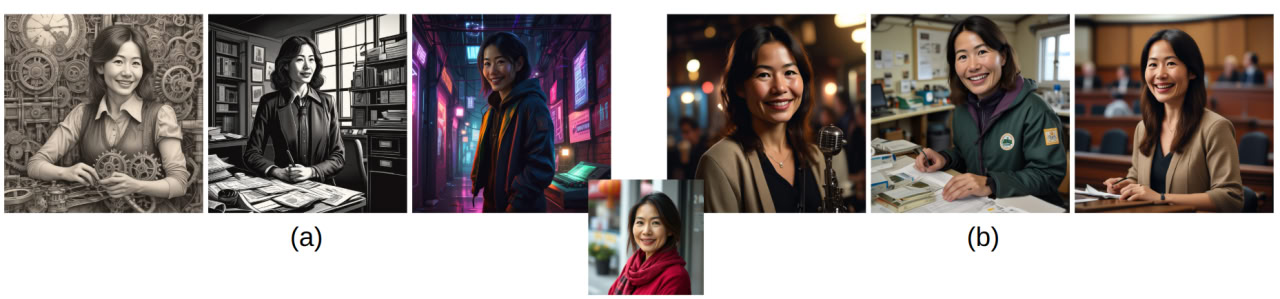}
  \caption{Different cases of user intention during ID-preserving generation: (a) - stylistic, (b) - realistic}
  \label{fig::generation_cases}
\end{figure}

 During training id-preserving adapters are proximally trained to reconstruct identity in the image, i.e. maximize similarity between the person in $c_{id}$ and $\hat x_0$ given conditional information. However, during inference, this is not strictly the case. Given pretrained ID-adapter we identify two common generation purposes -  \textit{stylistic} and \textit{realistic}. By "stylistic" we define generation that implies visual domain shift towards some priory known style, e.g. "pixel art" implies that generated image is expected to follow pixel-like visual appearance; by "realistic" setup we define generation that is not biased to any specific style or biased explicitly towards "realism" as specified in prompt. These cases correspond to different goals - in "stylistic" the user is less interested in facial features similarity and more in style following, while in realism situation is opposite, examples given in Figure \ref{fig::generation_cases}. In the following sections we develop two distinct approaches separately for these setups in the context of distilled diffusion models and unify them under one framework.

\raggedbottom

\subsection{Decoupled classifier free guidance}

\paragraph{Motivation} Firstly we revisit conventional classifier-free guidance (CFG) technique, widely used in diffusion models. It's definition is given in Equation 
\ref{eq::cfg}, $x_t$ is omitted for clarity. It can be seen that scale parameter $w$ impacts both conditioning strength on $c_{text}$ and $c_{id}$, not allowing any flexibility given two distinct conditions.

\begin{equation} \label{eq::cfg}
    \hat \epsilon = \epsilon(\varnothing, \varnothing) + w\cdot(\epsilon(c_{text}, c_{id}) - \epsilon(\varnothing, \varnothing))
\end{equation}

\paragraph{Basic formulation} Firstly we formulate baseline equation for decoupled classifier-free guidance (DCG) mechanism. Similar idea was applied before in instruct editing \cite{brooks2023instructpix2pix} to tune editability and fidelity. Similarly, we consider formulation of CFG in score functions and apply product rule to derive Equation \ref{eq::dcg-2}.

\begin{equation} \label{eq::dcg-2}
    \hat \epsilon = \epsilon(\varnothing, \varnothing) + \alpha\cdot(\epsilon(c_{id}, \varnothing) - \epsilon(\varnothing, \varnothing) + \beta \cdot (\epsilon(c_{text}, c_{id}) - \epsilon(c_{id}, \varnothing))
\end{equation} 

Detailed derivations and ablations of this and other possible variations of DCG are given in Appendix \ref{sec: dcg-variations-appendix}. In this setup $\alpha$ corresponds to strength of id conditioning and $\beta$ to textual strength conditioning, however it is not yet applicable to distilled models.

\begin{figure}[!htb]
  \centering
  \includegraphics[width=1.0\textwidth]{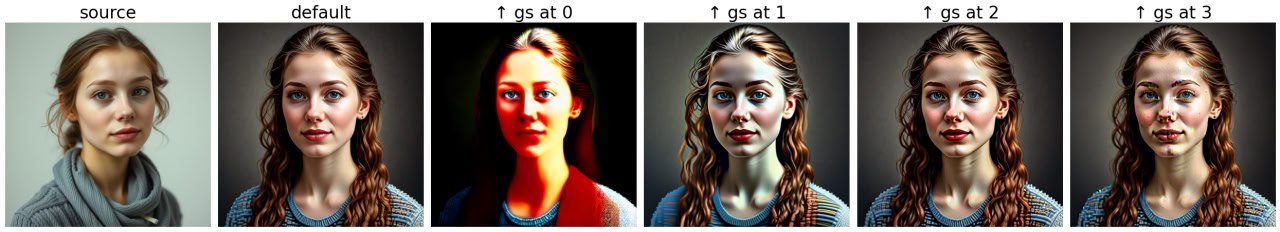}
  \caption{Scheduling effect on DCG, from right to left - baseline generation, single step alterations of $\alpha$ and $\beta$ coefficients to high value. In first steps image is completely corrupted, while last step introduces local visual artifacts} 
\label{fig: dcg_scaling_not_working}
\end{figure}

\paragraph{Making DCG work in few-steps} Distilled checkpoints of diffusion model out of the box are not suited for classifier guidance at all (exception in presented setup is LCM, which guidance scale can be tuned in range $[1.0, 2.0]$). Therefore we add two ideas to make it more stable. Firstly we ablate scheduling regimes of conventional classifier-free guidance for our setup. Contrary to findings in previous works (\cite{wang2024analysis}, \cite{starodubcev2024invertible}) we observe that applying any scheduling to both first and last steps of few-step models significantly alters resulting images, see Figure \ref{fig: dcg_scaling_not_working}, therefore we apply scheduling only during intermediate steps to minimize artifacts.

\raggedbottom

Secondly, we apply rescaling to DCG terms, as described in Eq. \ref{eq::dcg-rescaling}, which is inspired by rescaling trick introduced in \cite{lin2024common}. In first expression $\sigma_i$ and $\sigma_{ti}$ correspond to standard deviation of first and second terms of DCG. Second equation introduces interpolation trade-off between stability and quality, scaling hyper-parameter is generally fixed. This is a simple fix introduced with purpose to allow larger range of values of $\alpha$ and $\beta$ without introducing artifacts and works great in practice.

\begin{equation}\label{eq::dcg-rescaling}
    \begin{split}
        & \hat{\epsilon}_{\text{rescaled}} = \frac{\sigma_i + \sigma_{ti}}{2\sigma_{dcg}} \epsilon_{dcg},\\
        & \epsilon_{\text{final}} = \phi \cdot \hat{\epsilon}_{\text{rescaled}} + (1 - \phi) \epsilon_{dcg} \\
    \end{split}
\end{equation}

\begin{figure}[!htb]
  \centering
  \includegraphics[width=0.9\textwidth]{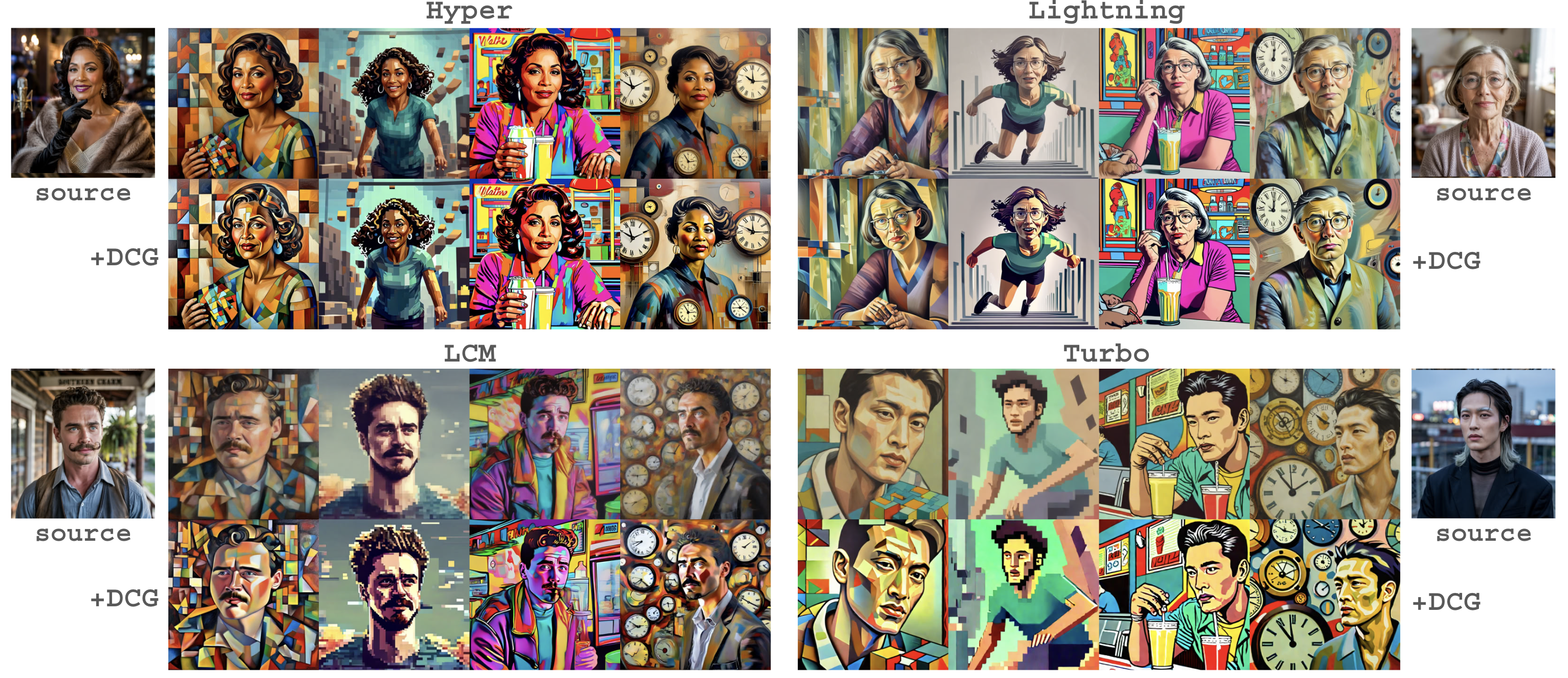}
  \caption{Visual result of applying DCG to stylistic generation with various models} 
\label{fig: dcg_result}
\end{figure}

\paragraph{Stylistic application} As a result of careful design for distilled models we tune DCG towards prompt following and image quality enhancement. We find it specifically useful in application with "style" prompts, as it enhances coherence of personalized generation with described style at low-level details, which can be observed in Figure \ref{fig: dcg_result}, without alteration of image structure. 

\raggedbottom

\subsection{Attention manipulation}

\begin{figure}[!htb]
  \centering
  \includegraphics[width=0.9\textwidth]{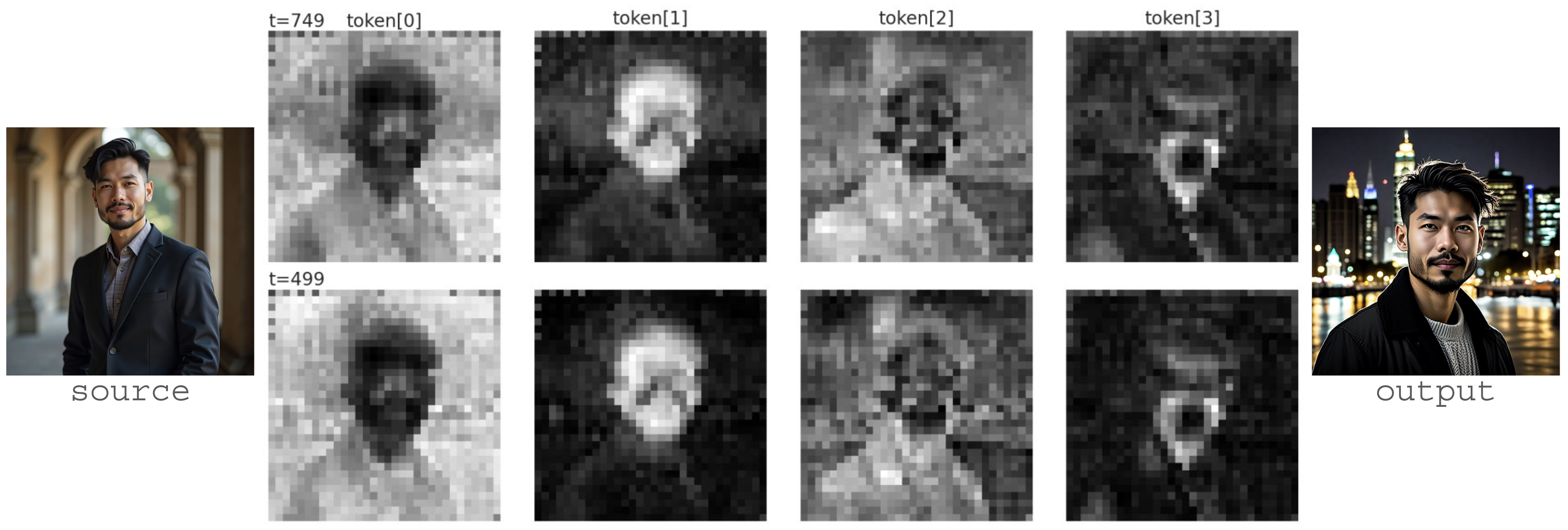}
  \caption{Visualization of attention maps timesteps 749 and 499 in decoupled block of SDXL in relation to generation output, specifically \texttt{up\_blocks.0.attentions.2.transformer\_blocks.6}} 
  \label{fig: dca-viz}
\end{figure}

\paragraph{Motivation} Attention maps in diffusion models are known to contain a lot of semantic and spatial information, which has been applied in numerous works of image editing (\cite{hertz2022prompt}, \cite{cao2023masactrl}, \cite{epstein2023diffusion}, \cite{titov2024guide}). Nuance of ID-adapters is that they train new cross-attention blocks within UNet to condition on visual information from $c_{id}$. We inspect these new blocks and visualize attention maps in Figure \ref{fig: dca-viz}  - it can bee seen that they share a lot of information with facial features and position in generated images, while also containing a lot of noisy signal about surrounding context, which can't be removed by changing \texttt{ip\_adapter\_scale} (see Eq. \ref{eq::adapterscale}). Therefore we opt to work with attention maps directly.

\raggedbottom

\paragraph{Basic formulation}  We begin with formulation of general Attention Manipulation (AM) algorithm in Equation \ref{eq::am_definition}. Main challenge is to construct such $f(\cdot): A \mapsto \tilde A$, where $A$ in attention map in decoupled blocks, that $\tilde A$ would allow achieve properties of 1) increasing face similarity/fidelity without significantly damaging prompt following 2) steering id-preserving generation towards more stable results, which we achieve by \emph{focusing attention on face regions}.
\begin{equation}\label{eq::am_definition}
        softmax(\frac{Q(K')^T}{\sqrt{d}}) \longrightarrow \color{red}f\color{black}(A) \longrightarrow \tilde{A}V' \longrightarrow z
\end{equation}

\textbf{Baseline scale-power transform} First transformation is designed via simple composition of scale and power transform applied to attention maps. Detailed ablation of this operations is given in Appendix \ref{sec: am-appendix}, in short power transformation applied to values less then 1 shifts everything closer to 0, while scaling linearly enhances attention mainly in meaningful tail of distribution with face region. 

\begin{equation}\label{eq::am_ps}
    f_{sp} := (\text{scale} ~\circ ~ \text{power})(A) = s \cdot A^p 
\end{equation}

\paragraph{Steering scheduled-softmask transform} Second transformation is designed in more tricky way to steer generation towards more stable, portrait-like images on average. This purpose is motivated by presence of "failure" cases, where for some reason id-preserving generation deviates towards unrealistic imagery or fails to preserve features in meaningful way, therefore requiring more global transformation, examples are given in Appendix \ref{sec: am-appendix}. It is constructed of following components 1) firstly Equation \ref{eq:am_softmask} performs an adaptive distribution shift of values less then $Q_p(A)$ towards 0 and others towards 1, strength of shift is defined by parameter $d$ 2) $d$ is scheduled to large value at first step to influence global structure of the image  3) smooth alignment with original attention statistics inspired by AdaIN \cite{huang2017arbitrary} is applied - normalizing transformed attention maps, modulate them using $\mu_A$ and $\sigma_A$ of original maps and interpolate between modulated and transformed versions, same operation is also applied to output of attention block (see Appendix \ref{sec: am-appendix}). Complete definition of $f_{ss}()$ is given in Equation \ref{eq:am_ss_final}. 

\begin{equation}\label{eq:am_softmask}
    \mathrm{softmask}(A, d, p) := s \cdot \sigma(\mathrm{norm}(\sigma(-d[ \mathrm{norm}(A) - Q_{p}(A) ])))
\end{equation}

\begin{equation}\label{eq:am_ss_final}
    f_{ss}(A) :=  w \cdot s \cdot \mathrm{softmask}(A,d,p) + (1-w) \cdot  \mathrm{AdaIN}(A, s \cdot \mathrm{softmask}(A,d,p))
\end{equation}

\begin{figure}[!htb]
    \centering
    \includegraphics[width=0.9\textwidth]{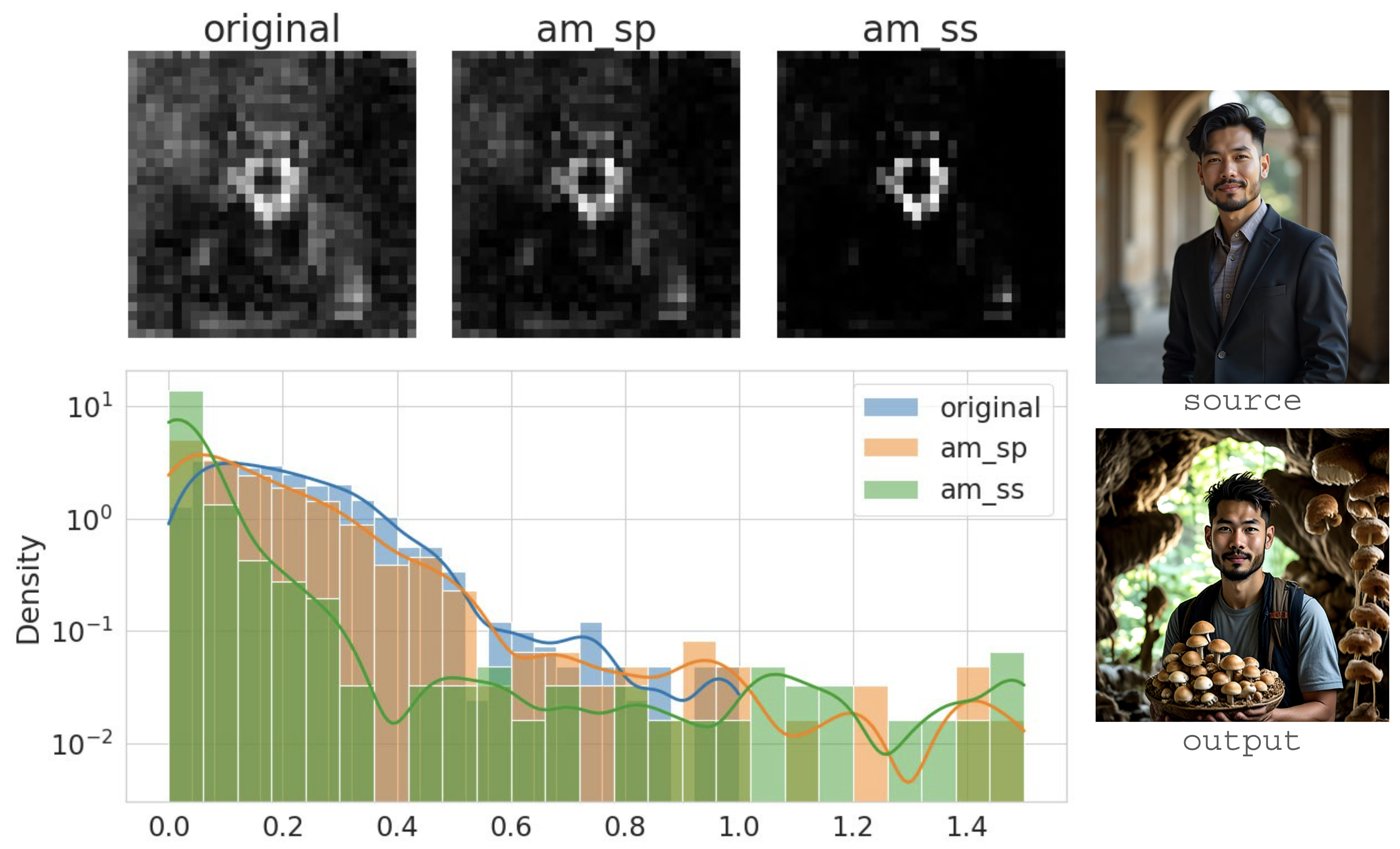}
    \caption{Visualizations of $f_{sp}$ and $f_{sm}$ transforms. At the top - visual result of transformation on the level of attention maps at certain block/step/token, bottom - distribution shift of attention values}
    \label{fig:transofrmed-attn-explained}
\end{figure}

In Equation \ref{eq:am_softmask} $norm(\cdot)$ denotes normalization and $Q_p(\cdot)$ is a p-th quintile function. In Figure \ref{fig:transofrmed-attn-explained} we visualize effect transforms have on attention values. We additionally study effect of transformation in terms two characteristics - similarity stability and face size distribution, results are presented in Appendix \ref{sec: am-appendix}. While both attentions are more focused on face regions, second transform achieves lower variance of face similarities without heavy tails and introduces more bias towards larger faces without any additional control. which can be seen in Figures \ref{fig: am_transforms_fulllora} and \ref{fig: am_transforms_lowlora}. 

\raggedbottom

\subsection{Full framework and evaluation}
\label{sec: framework-and-eval}

\begin{figure}[!htb]
  \centering
  \includegraphics[width=0.9\textwidth]{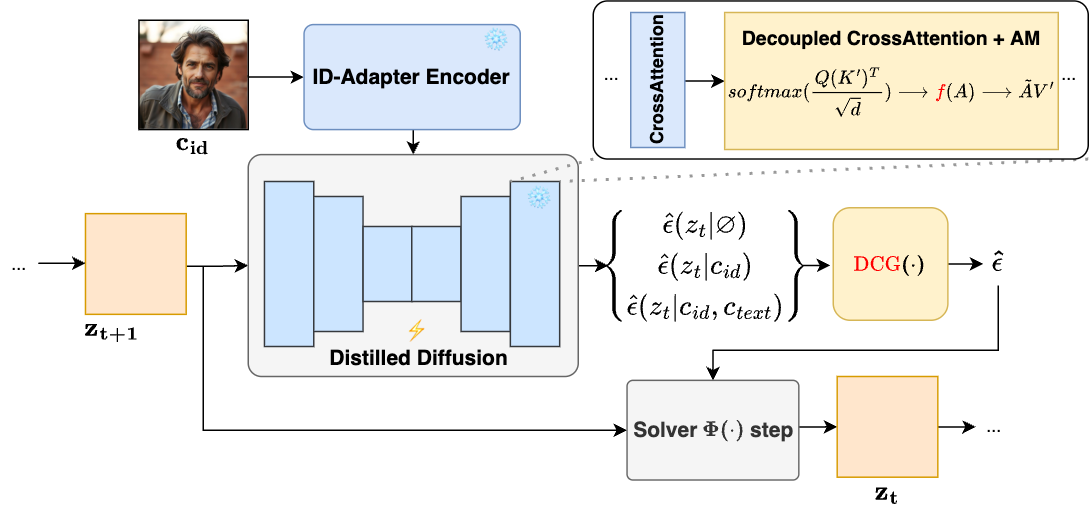}
  \caption{FastFace joint pipeline with proposed mechanisms - decoupled classifier free guidance, expanding on outputs of UNet, and attention manipulation as transform in decoupled blocks }
\label{fig::general_method_scheme}
\end{figure}

Together, presented mechanisms formulate joint framework of FastFace - through use of DCG and AM, which can be applied together or independently to any few-step sampling models, and are visualized in Figure \ref{fig::general_method_scheme}. In further sections we will demonstrate that these mechanisms work well together in general setting of id-preserving generation, as well as their respective setups of stylistic/realistic generations.

\raggedbottom

\paragraph{Necessity of novel evaluation} General motivation for new evaluation for ID-preserving generation comes from lack of one in the literature and understudied separation of ID-preservation use-cases. In previous works (\cite{ye2023ip}, \cite{wang2024instantid}, \cite{guo2024pulid}) authors rarely provide clarity about data which was used for evaluation, not allowing to fairly compare one method quality to another and understanding their strength and weaknesses. When evaluating these methods we find in a lot of cases they fail completely in practical setting, see Appendix \ref{sec: other-id-methods-analysis}. These issues can't be assessed without evaluation with transparent data, which is one of current work contributions.

\paragraph{Dataset details} Detailed description of collection and processing is given in Appendix 
\ref{sec: dataset-appendix}. In short, we collect a synthetic dataset of 54 high-resolution identity images from several models, ensuring diversity and filtering by mean similarity threshold withing identity groups. Prompts are constructed for two settings - 80 for realistic and 40 for stylistic. Product of these two sets is considered as full evaluation set, resulting in 2160 stylistic and 4320 realistic examples during evaluation.

\section{Experiments}

In further experiments we evaluate quality of baseline usage id-preserving adapter FaceID-Plus-v2 vs setups with additional mechanisms. We evaluate with Hyper, Lightning, LCM and Turbo checkpoints of SDXL in 4 step sampling regimes. In further sections for clear notation we denote scale-power transform as \texttt{AM1} and scheduled-softmask transform as \texttt{AM2}. All evaluations are done on A100 GPU.

\subsection{Metrics} \label{sec: metrics}

\paragraph{Common metrics} Metrics applied in both setups are face-similarity (ID), estimated as cosine distance between embeddings extracted by \texttt{buffalo-l} backbone from faces in source and generated images, and CLIP score (CLIP) between generated images and prompt computed with \texttt{clip/l-14} to estimate prompt alignment. We use LAION-Aesthetic (AE) reward model, which was trained on LAION subset, to estimate image quality/fidelity in general realistic image and in both in general evaluation and realistic subset \cite{aestheticpredictor}. Additionally we use ImageReward (IR) reward model to measure quality of stylistic images - as it was trained on synthetic data and is biased towards colorfulness and details and is more suited for that setting \cite{xu2023imagereward}.

\paragraph{Custom metrics} In stylistic setup we also rely on \texttt{face\_style\_score} (FSC) - CLIP-score calculated between cropped face and part of prompt describing style, measuring how well style is transferred on generated identity. In realistic setup we account for \texttt{face\_fail\_cnt} (FFC) - an integer metric which value represents amount of cases where detection model wasn't able to find any face in generated image.


\subsection{Quantitative results} \label{sec: exp-general}

We demonstrate effectiveness of FastFace framework by applying both DCG and AM together while varying \texttt{ip\_adapter\_scale}. Resulting fronts for Hyper model are displayed in Figure \ref{fig::hyper-fronts} (additional fronts for Lightning are reported in Appendix \ref{sec: full-pareto-fronts-lightning}), and is computed over joint set of realistic and stylistic setups (both mechanisms are applied at the same time). FastFace achieves alternative scaling behaviors of identity preservation w.r.t. other metrics, in most cases outperforming in both prompt following, identity preservation and quality, or introducing alternative trade offs. 

\begin{figure}[!htb]
  \centering  
    \subfloat[\centering \texttt{lora\_scale}=1.0]{\includegraphics[width=0.5\textwidth]{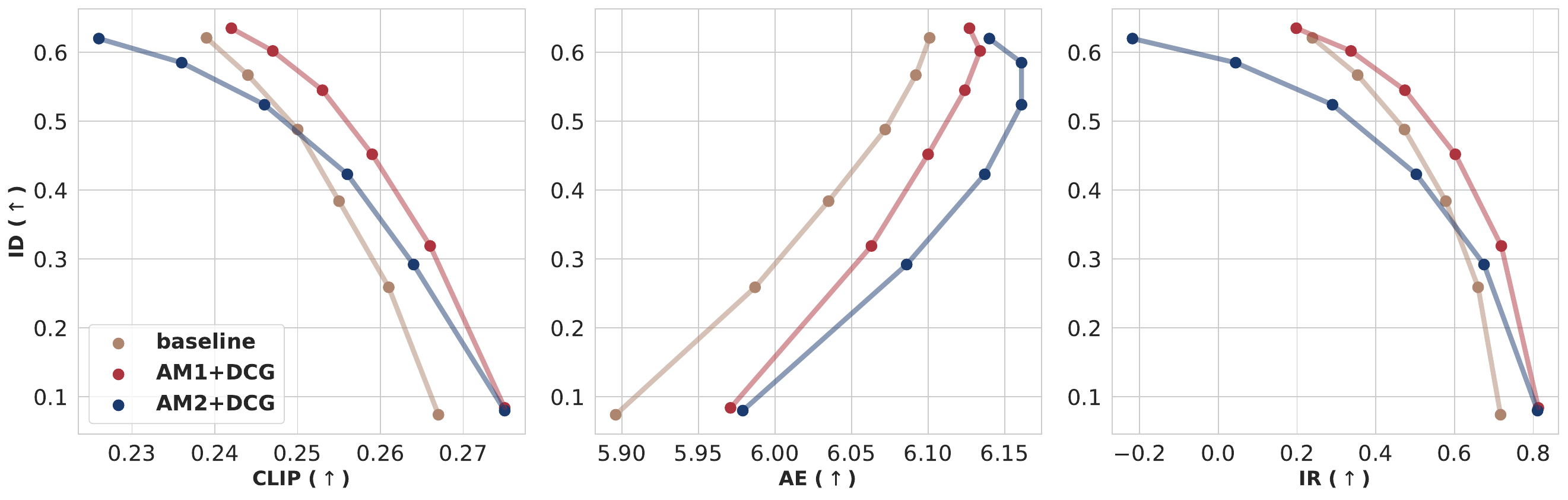}\label{fig:hyper-front-lora1}}
    \subfloat[\centering \texttt{lora\_scale}=0.5]{\includegraphics[width=0.5\textwidth]{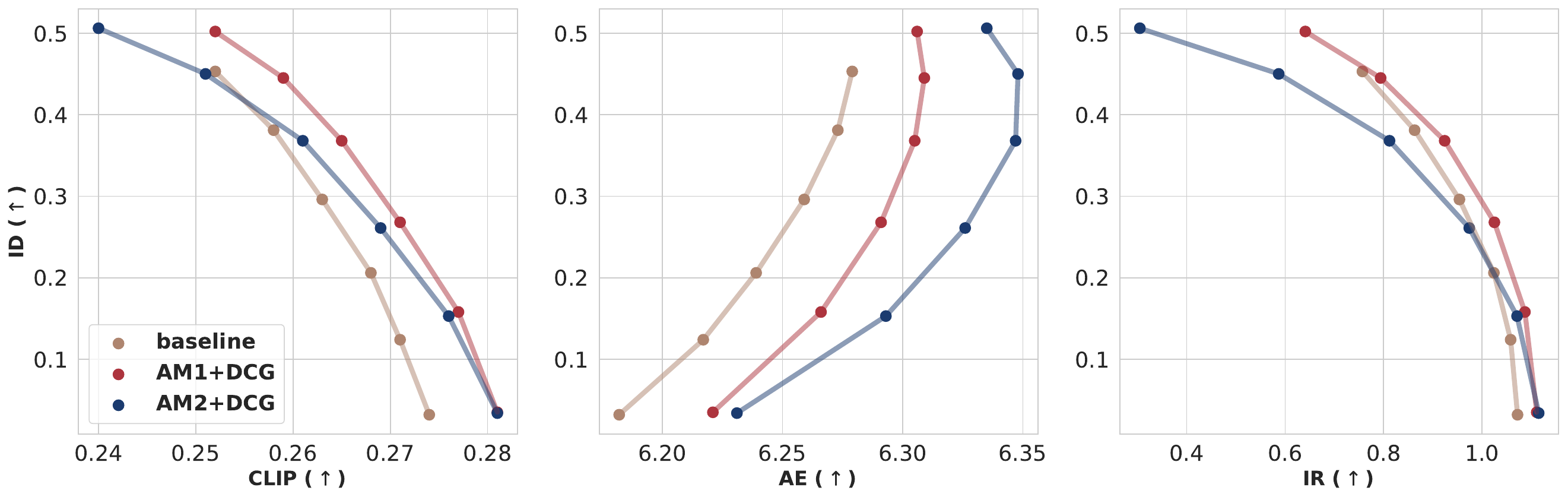}\label{fig:hyper-front-lora05}}
  \caption{Pareto fronts built for Hyper model metrics with different scales of LoRA}
  \label{fig::hyper-fronts}
\end{figure}
\raggedbottom

In Figure \ref{fig:hyper-front-lora05} and further sections we also analyze practical setting where \texttt{lora\_scale} assigned to LoRAs trained jointly with adapter module is lowered. This is a common practice in application of id-adapters to obtain more creative and natural generation outputs, but it also brings problem of instability, however FastFace is able to overcome it. We also report metric evaluation for fixed value of \texttt{ip\_adapter\_scale} across all models with full and lower LoRA scale in Table \ref{tab: full-table}. It is important to note that we do not tune mechanisms per checkpoint, all setups share default parameters.

\begin{table}[!h]
  \caption{Metric comparison of baseline setup against FastFace setups - $FF_{AM1}$ denotes application of DCG with scale-power transform, $FF_{AM2}$ - DCG with scheduled-softmask transform}
  \centering
  \begin{tabular}{@{}l cccc cccc @{}}
    \toprule
    \multirow{2}{*}{\textbf{Model}} &
    \multicolumn{4}{c}{\texttt{lora\_scale}$=1.0$} &
    \multicolumn{4}{c}{\texttt{lora\_scale}$=0.5$}\\
    \cmidrule(lr){2-5}\cmidrule(lr){6-9}
    & \texttt{ID} $\uparrow$ & CLIP $\uparrow$  & AE $\uparrow$ & FFC $\downarrow$ & \texttt{ID} $\uparrow$ & CLIP $\uparrow$  & AE $\uparrow$ & FFC $\downarrow$ \\
    \midrule
    Hyper (base) & 0.567  & \underline{0.244} & 6.092 &  2 & 0.381  & \underline{0.258} & 6.273 &  88  \\
    Hyper +  $FF_{AM1}$  & \textbf{0.602} & \textbf{0.247} & \underline{6.134}  & 2 & 0.445 & \textbf{0.259} & \underline{6.309} & \underline{72} \\
    Hyper +  $FF_{AM2}$  & \underline{0.585} & 0.236 & \textbf{6.161}  & \textbf{0} & \textbf{0.450} & 0.251 & \textbf{6.348}  & \textbf{34} \\
    \midrule 
    Lightning (base) & 0.504  & \underline{0.242} & 6.014 & 4 & 0.359  & 0.251 & 6.150 & 89 \\
    Lightning +  $FF_{AM1}$  & \textbf{0.543}  & \textbf{0.247} & \underline{6.112} & \underline{2} & 0.427  & \textbf{0.256} & \underline{6.254} & \underline{66}  \\
    Lightning +  $FF_{AM2}$  & \underline{0.542}  & 0.232 & \textbf{6.120} & \textbf{0} & \textbf{0.448}  & \underline{0.244} & \textbf{6.271} & \textbf{33} \\
    \midrule
    LCM (base) & 0.515 & \underline{0.243} & 5.770 & 53 & 0.344 & \textbf{0.258} & 5.911 & 288 \\
    LCM +  $FF_{AM1}$ & \underline{0.525}  & \textbf{0.244} & \underline{5.796} &  \underline{37} &  \underline{0.383} & \textbf{0.258} & \underline{5.968} & \underline{202} \\
    LCM +  $FF_{AM2}$ & \textbf{0.533}  & 0.229 & \textbf{5.807} & \textbf{20} & \textbf{0.406} & 0.246 & \textbf{5.979}  & \textbf{136} \\
    \midrule
    Turbo (base) & 0.336 & \underline{0.246} & 5.689 & 161 & 0.177 & \underline{0.257} & \underline{5.791} & \textbf{242} \\
    Turbo +  $FF_{AM1}$  & \textbf{0.416} & \textbf{0.249} & \underline{5.698} & \underline{139} & \underline{0.239} & \textbf{0.262}  & 5.757 & 431 \\
    Turbo +  $FF_{AM2}$  & \underline{0.409} & 0.242 & \textbf{5.707} & \textbf{94} & \textbf{0.244} & 0.256 &  \textbf{5.798}  & \underline{271} \\
    \bottomrule
  \end{tabular}
\label{tab: full-table}
\end{table}
\raggedbottom

\subsection{Qualitative results} 

\paragraph{Stylistic setup with DCG} In this experiment setup with stylistic prompts we apply only DCG formulated in Equation \ref{eq::dcg-2}. We schedule it to prioritize prompt following with values $\alpha(t) \in \{1.0, 1.5, 1.5, 1.0\}$ and $\beta(t) \in \{1.0, 3.0, 3.0, 1.0\}$, rescaling parameter is set $\phi=0.75$, fixed across models. In Figure \ref{fig: dcg_result} are given examples of effect DCG has on images generated with same conditions. Respective results in terms of metrics and fronts are given in Appendix \ref{sec: dcg-pareto-fronts-and-tables-appendix}.



\begin{figure}[!htb]
  \centering
  \subfloat[Hyper]{\includegraphics[width=0.5\textwidth]{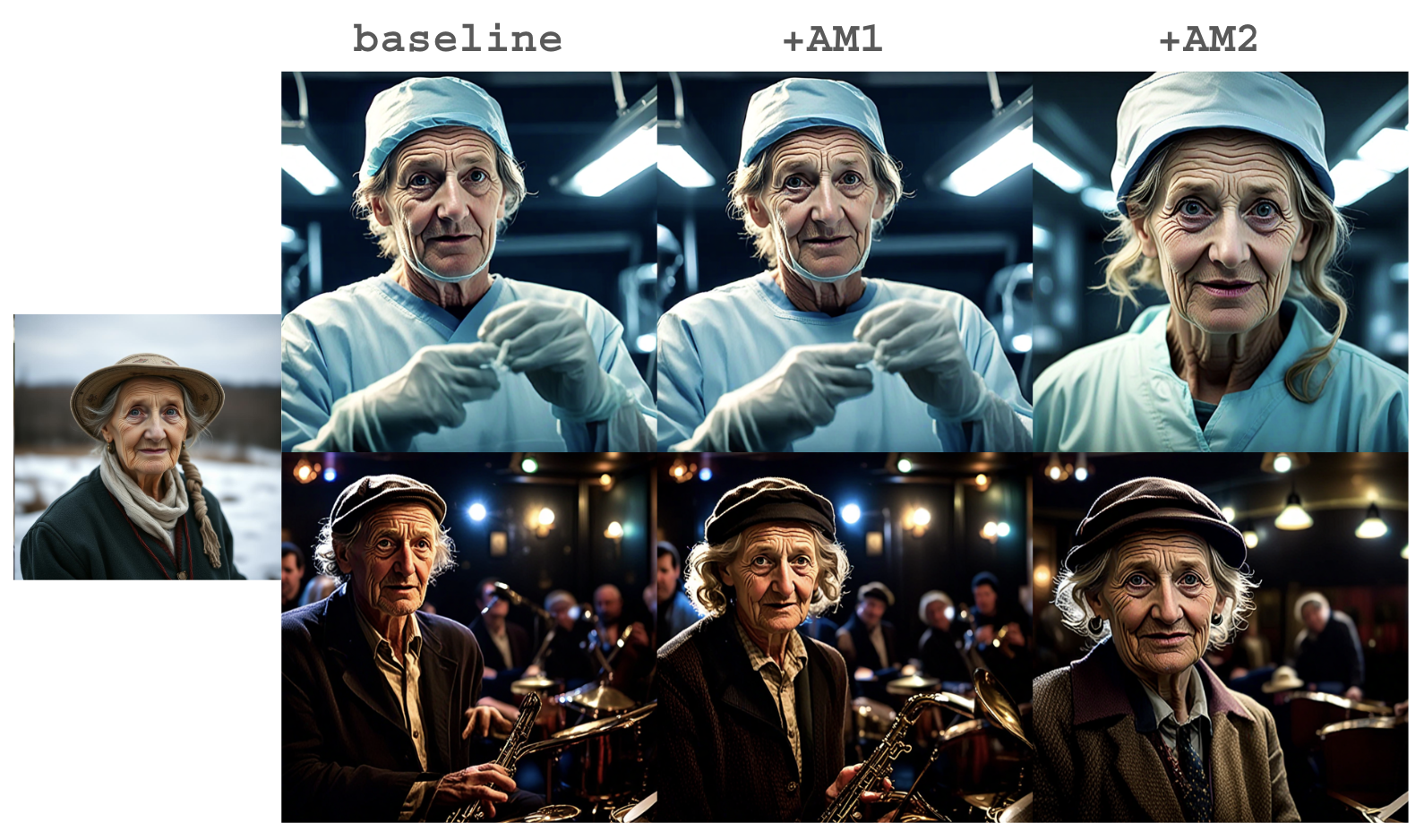}}
    \subfloat[\centering Lightning]{\includegraphics[width=0.5\textwidth]{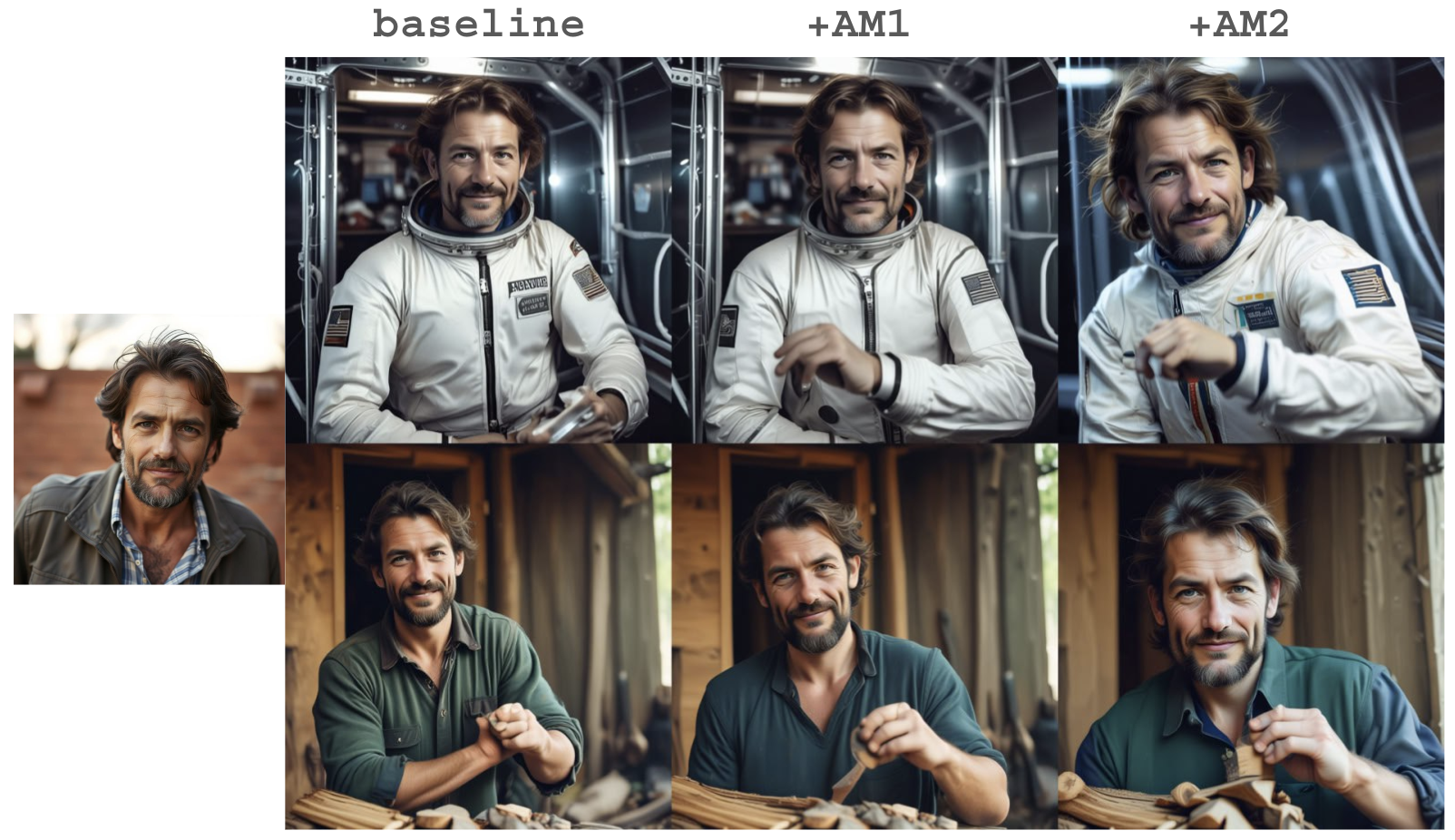}}
  \caption{Application of AM compared to baselines} 
\label{fig: am_transforms_fulllora}
\end{figure}
\begin{figure}[!htb]
  \centering
  \subfloat[Hyper]{\includegraphics[width=0.5\textwidth]{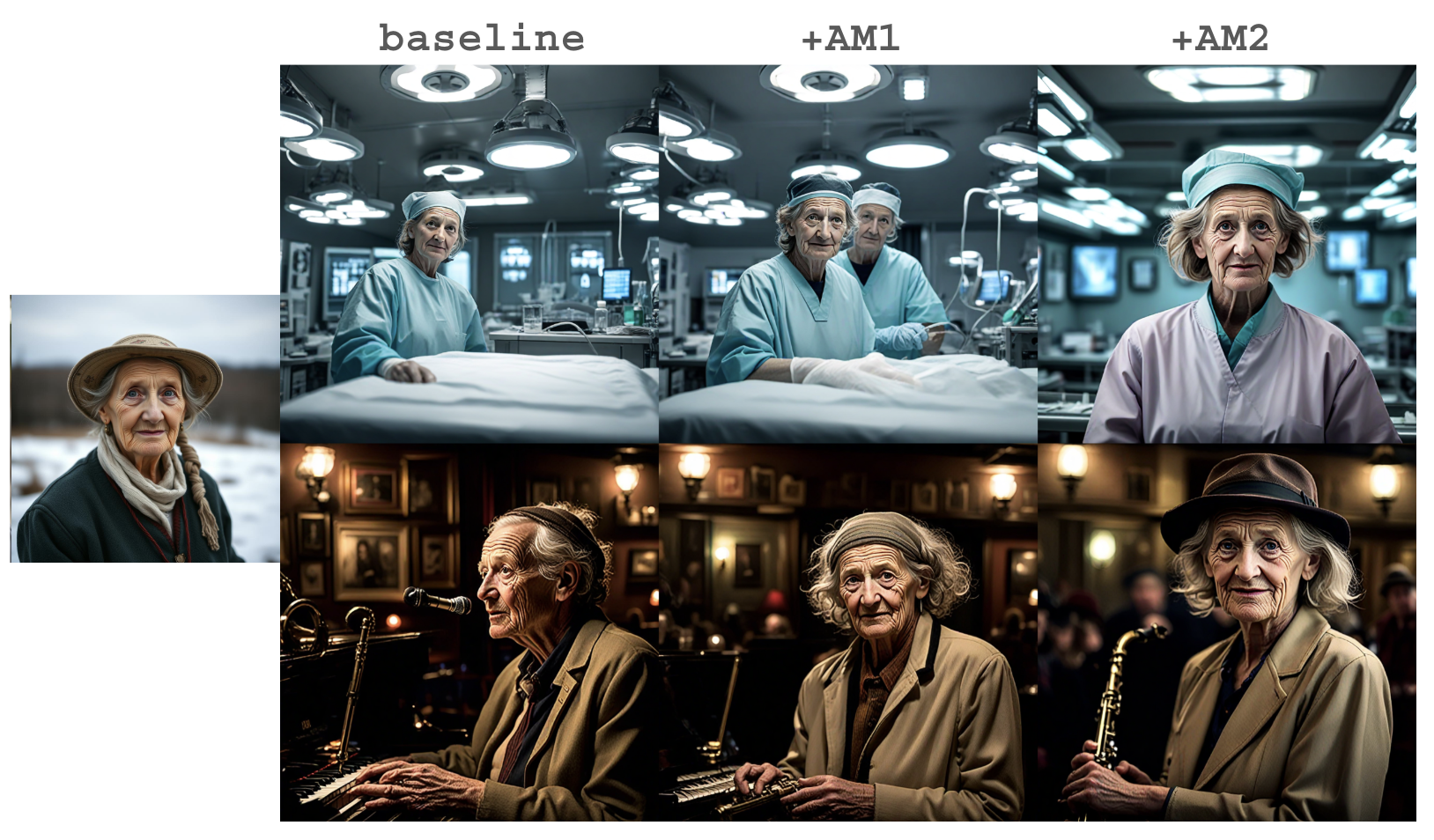}}
    \subfloat[\centering Lightning]{\includegraphics[width=0.5\textwidth]{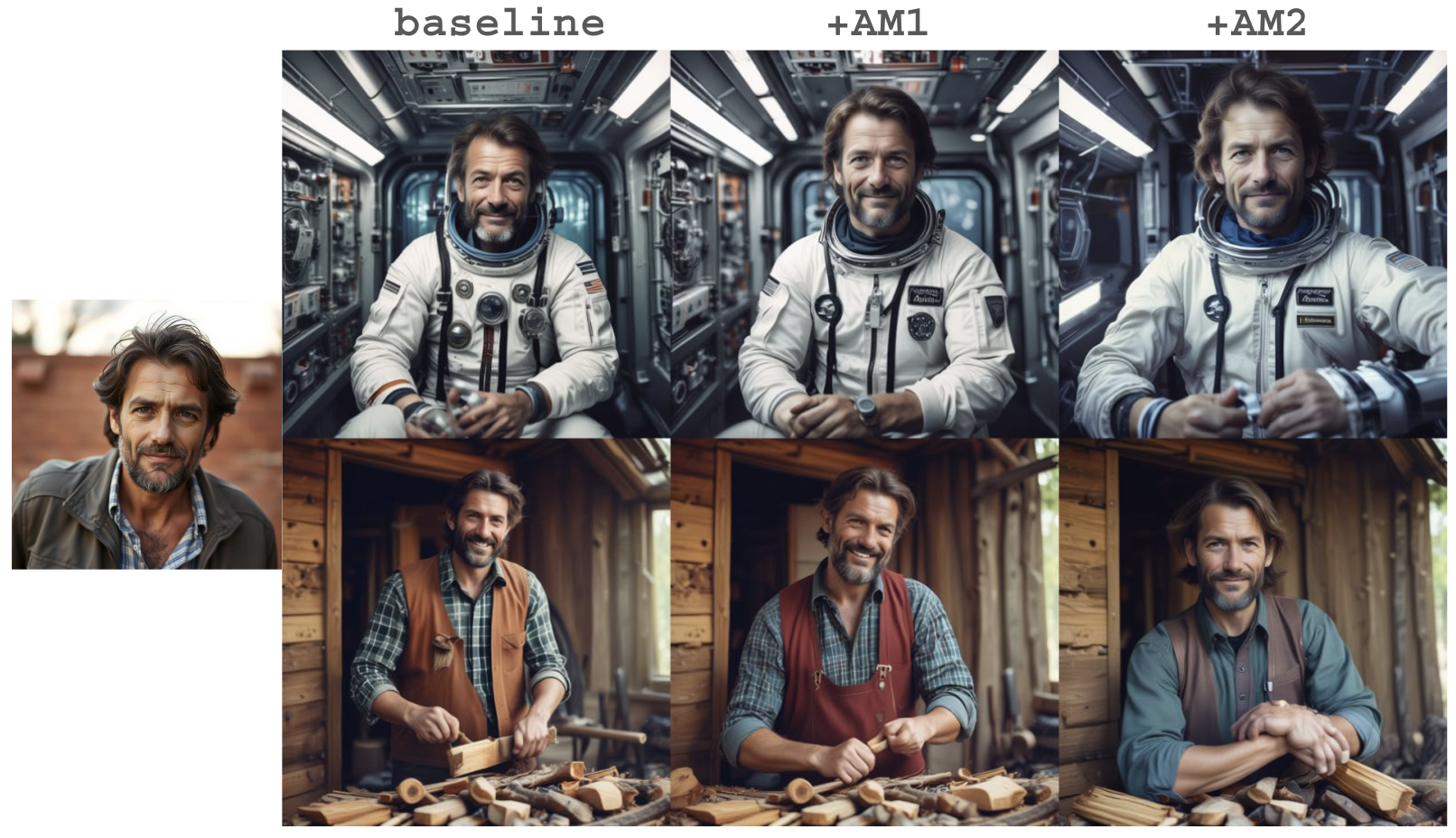}}
  \caption{Application of AM compared to baselines, \texttt{lora\_scale}=0.5} 
\label{fig: am_transforms_lowlora}
\end{figure}

\paragraph{Realistic setup with AM} In this experiment only $f_{sp}()$ and $f_{sm}()$ are applied to realistic evaluation set: detailed hyper-parameters, fronts and metrics are present in Appendix \ref{sec: am-pareto-fronts-and-tables-appendix}. It is demonstrated that proposed manipulations improve both ID and AE, while suffering minor loss in prompt following. Visual results for best checkpoints with fixed generation conditions are given in Figures \ref{fig: am_transforms_fulllora} and \ref{fig: am_transforms_lowlora}.

\raggedbottom

\raggedbottom

\section{Conclusion}

This work presents lightweight and easy-to-implement FastFace framework, which solves problem of adaptation of pretrained id-preserving generation adapter to distilled diffusion model without additional retraining. Included methods are developed for different cases of id-preserving generation - "stylistic", to better match style described in prompt, and "realistic", to enhance identity similarity or fidelity of the image. Presented contributions are evaluated in general, as well in specific scenarios on constructed evaluation dataset for id-preserving generation, showing generally better trade-offs in terms of identity preservation, prompt following and image quality.

\section{Limitations}
\label{sec: limitations}

Although proposed methods show promising results, scope of current work is limited to training-free methods, which are ultimately bottle-necked by distilled diffusion model checkpoint, and generally shows less impressive results in extreme cases such single-step sampling regime. It is a future work matter to address these limitation and adapt id-preserving generation to single-step models.

\newpage
\bibliographystyle{unsrt}  
\bibliography{references}     

\begin{thebibliography}{10}

\bibitem{ho2020ddpm}
Jonathan Ho, Ajay Jain, and Pieter Abbeel.
\newblock Denoising diffusion probabilistic models.
\newblock {\em NeurIPS}, 2020.

\bibitem{dhariwal2021diffusion}
Prafulla Dhariwal and Alexander Nichol.
\newblock Diffusion models beat gans on image synthesis.
\newblock {\em Advances in neural information processing systems}, 34:8780--8794, 2021.

\bibitem{rombach2022high}
Robin Rombach, Andreas Blattmann, Dominik Lorenz, Patrick Esser, and Bj{\"o}rn Ommer.
\newblock High-resolution image synthesis with latent diffusion models.
\newblock In {\em Proceedings of the IEEE/CVF conference on computer vision and pattern recognition}, pages 10684--10695, 2022.

\bibitem{podell2023sdxl}
Dustin Podell, Zion English, Kyle Lacey, Andreas Blattmann, Tim Dockhorn, Jonas M{\"u}ller, Joe Penna, and Robin Rombach.
\newblock Sdxl: Improving latent diffusion models for high-resolution image synthesis.
\newblock {\em arXiv preprint arXiv:2307.01952}, 2023.

\bibitem{esser2024scaling}
Patrick Esser, Sumith Kulal, Andreas Blattmann, Rahim Entezari, Jonas M{\"u}ller, Harry Saini, Yam Levi, Dominik Lorenz, Axel Sauer, Frederic Boesel, et~al.
\newblock Scaling rectified flow transformers for high-resolution image synthesis.
\newblock In {\em Forty-first international conference on machine learning}, 2024.

\bibitem{flux2024}
Black~Forest Labs.
\newblock Flux.
\newblock \url{https://github.com/black-forest-labs/flux}, 2024.

\bibitem{luo2023latent}
Simian Luo, Yiqin Tan, Longbo Huang, Jian Li, and Hang Zhao.
\newblock Latent consistency models: Synthesizing high-resolution images with few-step inference.
\newblock {\em arXiv preprint arXiv:2310.04378}, 2023.

\bibitem{sauer2024fast}
Axel Sauer, Frederic Boesel, Tim Dockhorn, Andreas Blattmann, Patrick Esser, and Robin Rombach.
\newblock Fast high-resolution image synthesis with latent adversarial diffusion distillation.
\newblock In {\em SIGGRAPH Asia 2024 Conference Papers}, pages 1--11, 2024.

\bibitem{lin2024sdxl}
Shanchuan Lin, Anran Wang, and Xiao Yang.
\newblock Sdxl-lightning: Progressive adversarial diffusion distillation.
\newblock {\em arXiv preprint arXiv:2402.13929}, 2024.

\bibitem{ren2024hyper}
Yuxi Ren, Xin Xia, Yanzuo Lu, Jiacheng Zhang, Jie Wu, Pan Xie, Xing Wang, and Xuefeng Xiao.
\newblock Hyper-sd: Trajectory segmented consistency model for efficient image synthesis.
\newblock {\em arXiv preprint arXiv:2404.13686}, 2024.

\bibitem{sauer2024adversarial}
Axel Sauer, Dominik Lorenz, Andreas Blattmann, and Robin Rombach.
\newblock Adversarial diffusion distillation.
\newblock In {\em European Conference on Computer Vision}, pages 87--103. Springer, 2024.

\bibitem{ye2023ip}
Hu~Ye, Jun Zhang, Sibo Liu, Xiao Han, and Wei Yang.
\newblock Ip-adapter: Text compatible image prompt adapter for text-to-image diffusion models.
\newblock {\em arXiv preprint arXiv:2308.06721}, 2023.

\bibitem{li2024photomaker}
Zhen Li, Mingdeng Cao, Xintao Wang, Zhongang Qi, Ming-Ming Cheng, and Ying Shan.
\newblock Photomaker: Customizing realistic human photos via stacked id embedding.
\newblock In {\em Proceedings of the IEEE/CVF conference on computer vision and pattern recognition}, pages 8640--8650, 2024.

\bibitem{wang2024instantid}
Qixun Wang, Xu~Bai, Haofan Wang, Zekui Qin, Anthony Chen, Huaxia Li, Xu~Tang, and Yao Hu.
\newblock Instantid: Zero-shot identity-preserving generation in seconds.
\newblock {\em arXiv preprint arXiv:2401.07519}, 2024.

\bibitem{guo2024pulid}
Zinan Guo, Yanze Wu, Chen Zhuowei, Peng Zhang, Qian He, et~al.
\newblock Pulid: Pure and lightning id customization via contrastive alignment.
\newblock {\em Advances in neural information processing systems}, 37:36777--36804, 2024.

\bibitem{jiang2025infiniteyou}
Liming Jiang, Qing Yan, Yumin Jia, Zichuan Liu, Hao Kang, and Xin Lu.
\newblock Infiniteyou: Flexible photo recrafting while preserving your identity.
\newblock {\em arXiv preprint arXiv:2503.16418}, 2025.

\bibitem{zhang2023adding}
Lvmin Zhang, Anyi Rao, and Maneesh Agrawala.
\newblock Adding conditional control to text-to-image diffusion models.
\newblock In {\em Proceedings of the IEEE/CVF international conference on computer vision}, pages 3836--3847, 2023.

\bibitem{xiao2023ccm}
Jie Xiao, Kai Zhu, Han Zhang, Zhiheng Liu, Yujun Shen, Yu~Liu, Xueyang Fu, and Zheng-Jun Zha.
\newblock Ccm: Adding conditional controls to text-to-image consistency models.
\newblock {\em arXiv preprint arXiv:2312.06971}, 2023.

\bibitem{parmar2024one}
Gaurav Parmar, Taesung Park, Srinivasa Narasimhan, and Jun-Yan Zhu.
\newblock One-step image translation with text-to-image models.
\newblock {\em arXiv preprint arXiv:2403.12036}, 2024.

\bibitem{lin2024ctrl}
Han Lin, Jaemin Cho, Abhay Zala, and Mohit Bansal.
\newblock Ctrl-adapter: An efficient and versatile framework for adapting diverse controls to any diffusion model.
\newblock {\em arXiv preprint arXiv:2404.09967}, 2024.

\bibitem{kolors}
Kolors Team.
\newblock Kolors: Effective training of diffusion model for photorealistic text-to-image synthesis.
\newblock {\em arXiv preprint}, 2024.

\bibitem{ruiz2023dreambooth}
Nataniel Ruiz, Yuanzhen Li, Varun Jampani, Yael Pritch, Michael Rubinstein, and Kfir Aberman.
\newblock Dreambooth: Fine tuning text-to-image diffusion models for subject-driven generation.
\newblock In {\em Proceedings of the IEEE/CVF conference on computer vision and pattern recognition}, pages 22500--22510, 2023.

\bibitem{salimans2022progressive}
Tim Salimans and Jonathan Ho.
\newblock Progressive distillation for fast sampling of diffusion models.
\newblock {\em arXiv preprint arXiv:2202.00512}, 2022.

\bibitem{song2023consistency}
Yang Song, Prafulla Dhariwal, Mark Chen, and Ilya Sutskever.
\newblock Consistency models.
\newblock 2023.

\bibitem{yin2024one}
Tianwei Yin, Micha{\"e}l Gharbi, Richard Zhang, Eli Shechtman, Fredo Durand, William~T Freeman, and Taesung Park.
\newblock One-step diffusion with distribution matching distillation.
\newblock In {\em Proceedings of the IEEE/CVF conference on computer vision and pattern recognition}, pages 6613--6623, 2024.

\bibitem{ke2023repurposing}
Bingxin Ke, Anton Obukhov, Shengyu Huang, Nando Metzger, Rodrigo~Caye Daudt, and Konrad Schindler.
\newblock Repurposing diffusion-based image generators for monocular depth estimation.
\newblock In {\em Proceedings of the IEEE/CVF Conference on Computer Vision and Pattern Recognition (CVPR)}, 2024.

\bibitem{chen2024pixart}
Junsong Chen, Yue Wu, Simian Luo, Enze Xie, Sayak Paul, Ping Luo, Hang Zhao, and Zhenguo Li.
\newblock Pixart-$\{$$\backslash$delta$\}$: Fast and controllable image generation with latent consistency models.
\newblock {\em arXiv preprint arXiv:2401.05252}, 2024.

\bibitem{xu2024ctrlora}
Yifeng Xu, Zhenliang He, Shiguang Shan, and Xilin Chen.
\newblock Ctrlora: An extensible and efficient framework for controllable image generation.
\newblock {\em arXiv preprint arXiv:2410.09400}, 2024.

\bibitem{brooks2023instructpix2pix}
Tim Brooks, Aleksander Holynski, and Alexei~A Efros.
\newblock Instructpix2pix: Learning to follow image editing instructions.
\newblock In {\em Proceedings of the IEEE/CVF conference on computer vision and pattern recognition}, pages 18392--18402, 2023.

\bibitem{wang2024analysis}
Xi~Wang, Nicolas Dufour, Nefeli Andreou, Marie-Paule Cani, Victoria~Fern{\'a}ndez Abrevaya, David Picard, and Vicky Kalogeiton.
\newblock Analysis of classifier-free guidance weight schedulers.
\newblock {\em arXiv preprint arXiv:2404.13040}, 2024.

\bibitem{starodubcev2024invertible}
Nikita Starodubcev, Mikhail Khoroshikh, Artem Babenko, and Dmitry Baranchuk.
\newblock Invertible consistency distillation for text-guided image editing in around 7 steps.
\newblock {\em arXiv preprint arXiv:2406.14539}, 2024.

\bibitem{lin2024common}
Shanchuan Lin, Bingchen Liu, Jiashi Li, and Xiao Yang.
\newblock Common diffusion noise schedules and sample steps are flawed.
\newblock In {\em Proceedings of the IEEE/CVF winter conference on applications of computer vision}, pages 5404--5411, 2024.

\bibitem{hertz2022prompt}
Amir Hertz, Ron Mokady, Jay Tenenbaum, Kfir Aberman, Yael Pritch, and Daniel Cohen-Or.
\newblock Prompt-to-prompt image editing with cross attention control.
\newblock {\em arXiv preprint arXiv:2208.01626}, 2022.

\bibitem{cao2023masactrl}
Mingdeng Cao, Xintao Wang, Zhongang Qi, Ying Shan, Xiaohu Qie, and Yinqiang Zheng.
\newblock Masactrl: Tuning-free mutual self-attention control for consistent image synthesis and editing.
\newblock In {\em Proceedings of the IEEE/CVF international conference on computer vision}, pages 22560--22570, 2023.

\bibitem{epstein2023diffusion}
Dave Epstein, Allan Jabri, Ben Poole, Alexei Efros, and Aleksander Holynski.
\newblock Diffusion self-guidance for controllable image generation.
\newblock {\em Advances in Neural Information Processing Systems}, 36:16222--16239, 2023.

\bibitem{titov2024guide}
Vadim Titov, Madina Khalmatova, Alexandra Ivanova, Dmitry Vetrov, and Aibek Alanov.
\newblock Guide-and-rescale: Self-guidance mechanism for effective tuning-free real image editing.
\newblock In {\em European Conference on Computer Vision}, pages 235--251. Springer, 2024.

\bibitem{huang2017arbitrary}
Xun Huang and Serge Belongie.
\newblock Arbitrary style transfer in real-time with adaptive instance normalization.
\newblock In {\em Proceedings of the IEEE international conference on computer vision}, pages 1501--1510, 2017.

\bibitem{aestheticpredictor}
LAION.
\newblock {Aesthetic model predictor - GitHub repository}.
\newblock \url{https://github.com/LAION-AI/aesthetic-predictor}, 2022.

\bibitem{xu2023imagereward}
Jiazheng Xu, Xiao Liu, Yuchen Wu, Yuxuan Tong, Qinkai Li, Ming Ding, Jie Tang, and Yuxiao Dong.
\newblock Imagereward: Learning and evaluating human preferences for text-to-image generation.
\newblock {\em Advances in Neural Information Processing Systems}, 36:15903--15935, 2023.

\bibitem{insightface}
Deep Insight.
\newblock Insightface: 2d and 3d face analysis project.
\newblock \url{https://github.com/deepinsight/insightface}, 2023.

\bibitem{song2020score}
Yang Song, Jascha Sohl-Dickstein, Diederik~P Kingma, Abhishek Kumar, Stefano Ermon, and Ben Poole.
\newblock Score-based generative modeling through stochastic differential equations.
\newblock {\em arXiv preprint arXiv:2011.13456}, 2020.

\end{thebibliography}

\newpage
\appendix

\section{Technical Appendices and Supplementary Material}

\subsection{Details of evaluation dataset}
\label{sec: dataset-appendix}

\begin{figure}[!htb]
  \centering
  \includegraphics[width=0.9\textwidth]{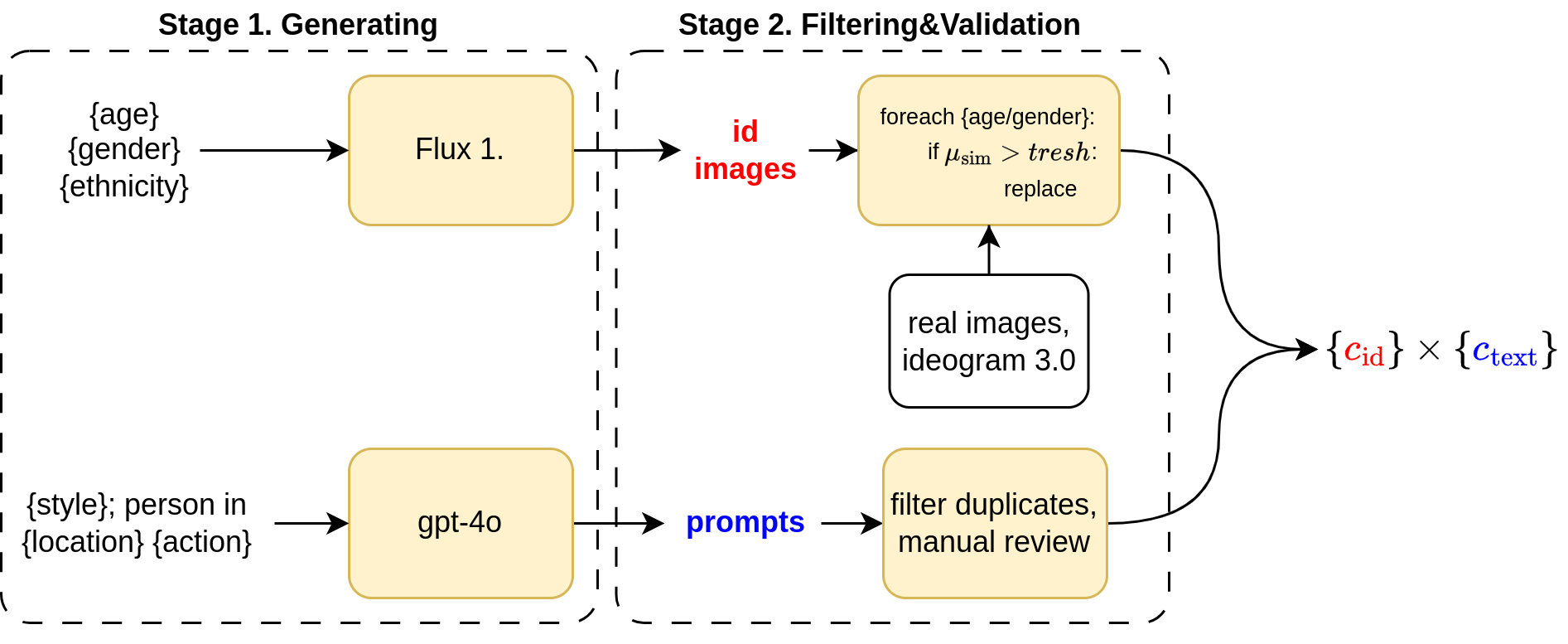}
  \caption{Evaluation dataset preparation pipeline}
\label{fig: dataset_build}
\end{figure}

\begin{figure}[!htb]
  \centering
  \includegraphics[width=0.8\textwidth]{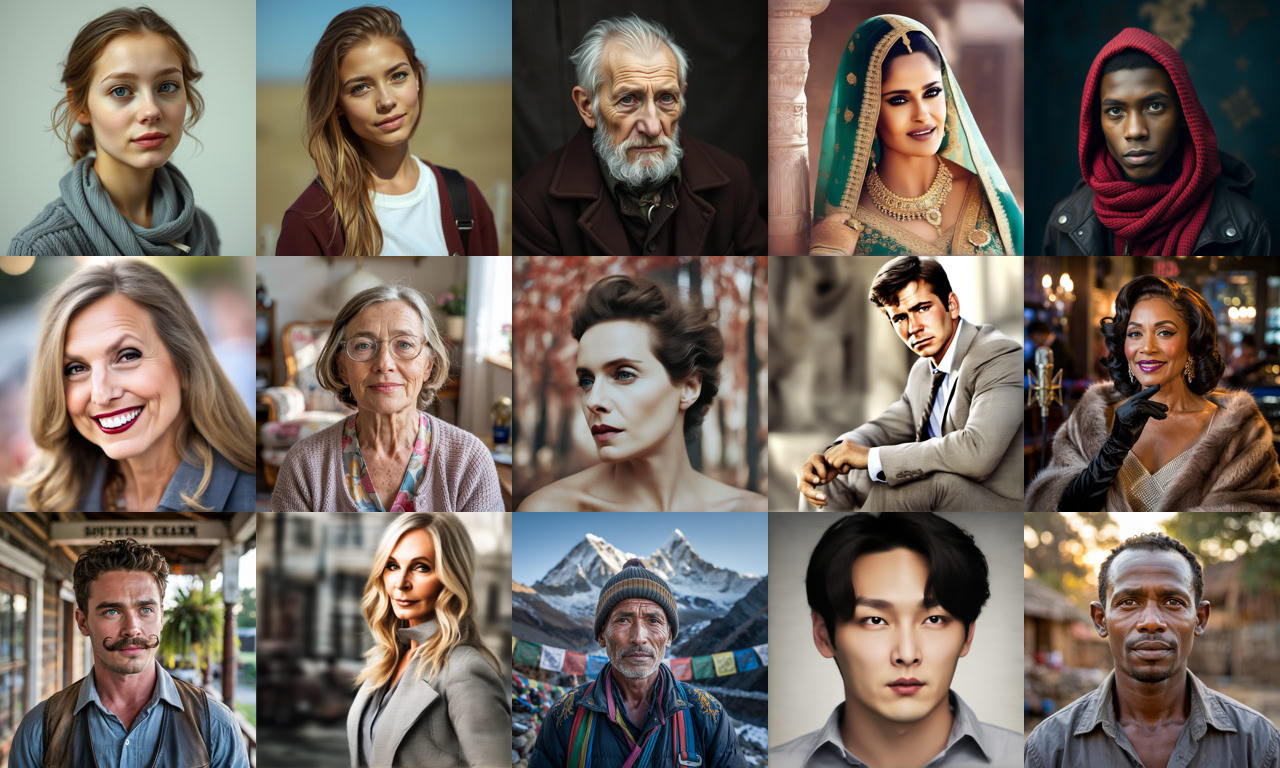}
  \caption{Evaluation dataset identity samples}
\label{fig: dataset_samples}
\end{figure}

We develop an evaluation dataset consisting of 54 high quality identity images and 120 prompts, which are used as input conditions for generation and further evaluation. Identity images are synthetic images from models such as Flux and Ideogram 3.0 (\cite{flux2024}), representing different age groups (young, middle age and old), genders and ethnicities, examples are presented in Figure \ref{fig: dataset_samples}. Part of images was also synthesized using id-preserving methods with from real identities, thus avoiding bias towards only synthetic facial features. Additionally, to ensure variance within groups of identities of same gender and age, further cleaning was done by thresholding and replacing identity images with largest mean face similarity to others, i.e. if $
\frac{1}{n-1}\Sigma_{j, j \neq i} sim(c_{i}, c_{j}) > 0.3$ for $c_{i}$ within group, it was discarded. Prompt description were also synthetically generated using Chat-GPT version of November 2024, generally following structure of \texttt{style + ';' + 'Person' + location + action}, and then additionally cleared and enriched. Prompts are categorized into two groups - 80 "realistic" prompts and 40 "style" prompts with certain style. Product of id images and prompts from category is considered as evaluation set, resulting in two sets - stylistic with 2160 and realistic with 4320 examples. Schematic depiction of the data collection is visualized in Figure \ref{fig: dataset_build}.

\raggedbottom

\subsection{Dataset samples}
\label{sec: metrics-appendix}

\raggedbottom


\subsection{Certain ID-methods inference failure examples}
\label{sec: other-id-methods-analysis}

Below we provide examples of recent id-preserving generation methods that we found to have limitations in terms of application with our evaluation set.

\begin{figure}[!htb]
  \centering
  \includegraphics[width=0.8\textwidth]{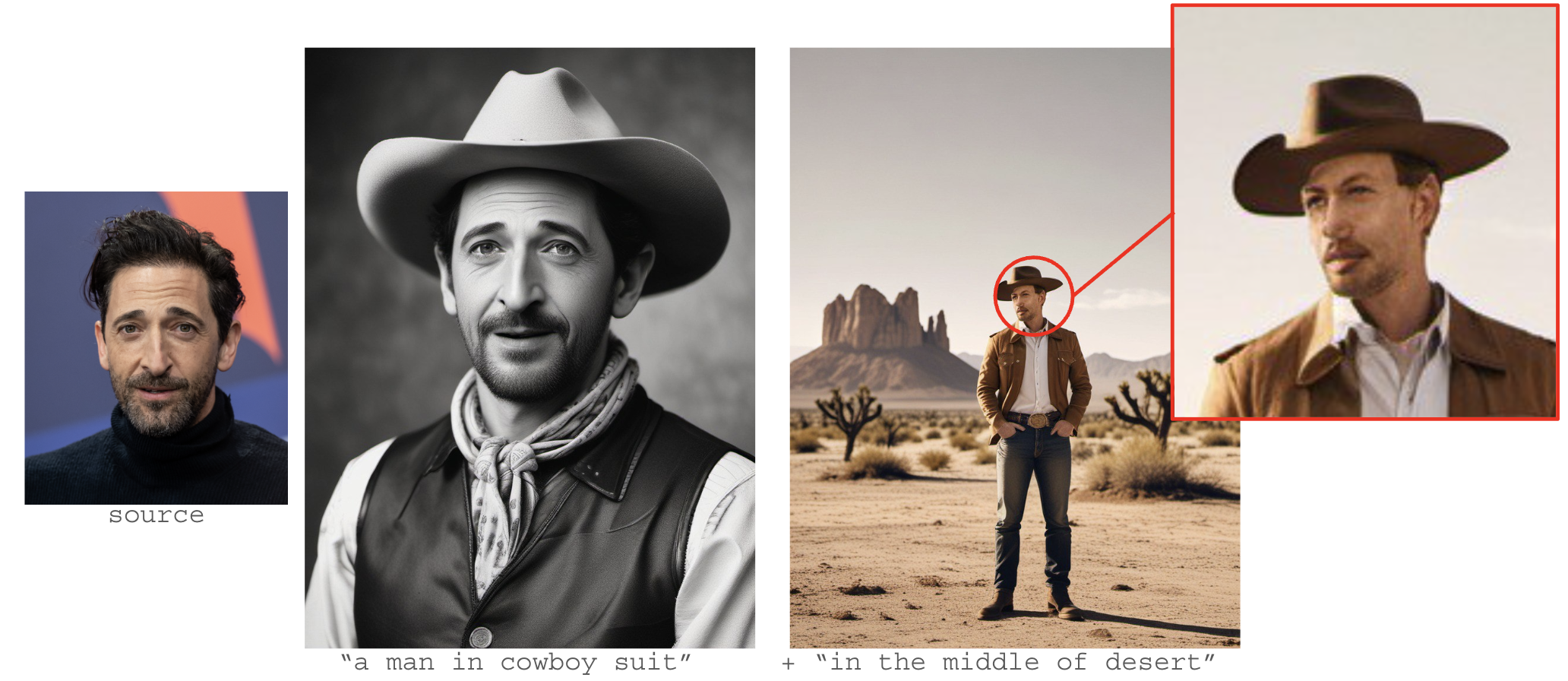}
  \caption{Demonstration of common case of failure for PuLID method - method lacks bias to human-centric generation to perform identity preservation, especially with small faces.} 
  \label{fig: pulid_failure}
\end{figure}

\paragraph{PuLID} In Figure \ref{fig: pulid_failure} we provide example common failure for PulID method. From our experiments we find that it is not applicable with prompts that have description of context like location and action, which our evaluation set prompts have. We hypothesize that this effect is rooted in aligned training of PuLID, where inner representations of UNet are regularized to match generation without $c_{id}$ condition - in our experiments we found that in baseline setup FFC metric accounts around for 50\% of sampled images failing (meaning around half of images doesn't have any identity detected).

\raggedbottom

\begin{figure}[!htb]
  \centering
  \includegraphics[width=0.8\textwidth]{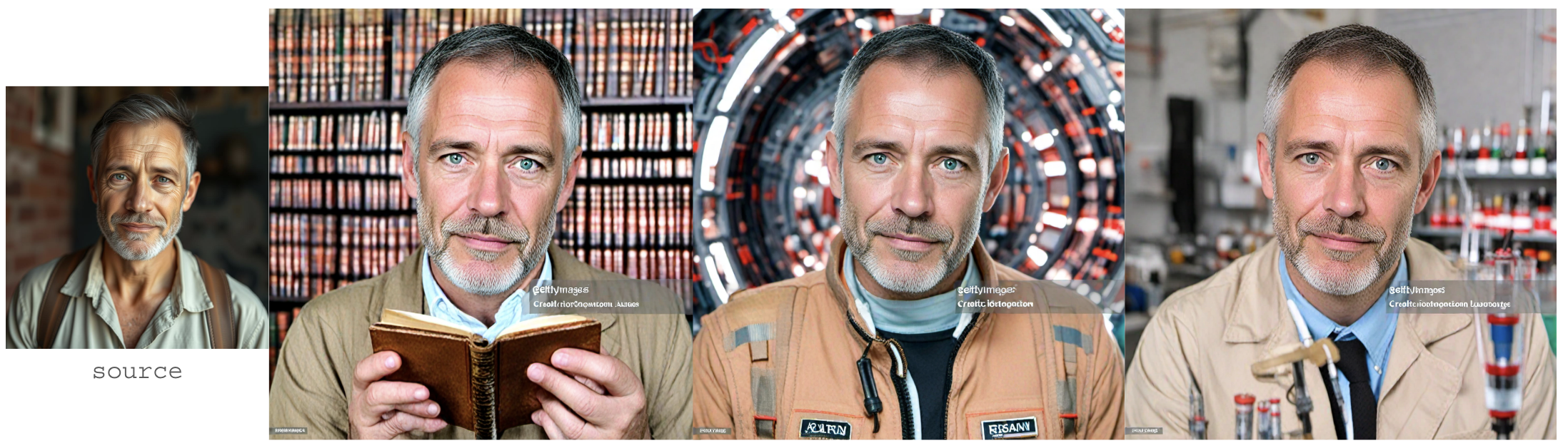}
  \caption{Demonstration of common case of failure for InstantID method - generated images are highly constrained and often omit details in the prompt, prompts used for generation: \textit{"Person in an ancient library reading"}, \textit{"Person in a futuristic space station repairing equipment"}, \textit{"Person in a high-tech laboratory conducting experiments"}} 
  \label{fig: instantid_failure}
\end{figure}

\paragraph{InstantID} This method is example of opposite behavior - it's pipeline includes ControlNet-like module that is conditioned on face key-points, which are extracted from source image by standard CV packages (e.g. \texttt{insightface} \cite{insightface}). However, when tested against multiple different prompts, we observe in Fig. 
\ref{fig: instantid_failure} that despite showing state of the art in terms of face preservation, outperforming any other method, it lacks prompt following and variability, not being able to properly follow details regarding background and person body position (additionally it has large bias towards watermark generation with 1:1 resolutions).

\raggedbottom

\subsection{DCG variations and derivations}
\label{sec: dcg-variations-appendix}

\paragraph{Preliminary} To simplify derivation process let's recall that reverse diffusion process is formulated in terms of score function $\nabla_{x_t}\log p(x_t|y)$ \cite{song2020score}, where $x_t$ is noised latent and $y$ is conditional information, in text2image models being prompt. Then classifier guidance can be derived as below, where in Eq. \ref{eq:cg} $w$ is added as a hyper-parameter to control conditioning strength. 

\begin{align}
    \nabla_{x_t}\log p(x_t|y) &= \nabla_{x_t}\log(\frac{p(y| x_t)p(x_t)}{p(y)}) \\
     &= \nabla_{x_t}\log p(y|x_t) + \nabla_{x_t} \log p(x_t)  - \nabla_{x_t}\log p(y) \\
     &\Rightarrow \nabla_{x_t} \log p(x_t) + w\cdot \nabla_{x_t}\log p(y|x_t) \label{eq:cg}
\end{align}

Then to arrive to classifier-free guidance (which removes need for learning classifier $f(y|x_t)$ for estimation of $\nabla_{x_t}\log p(y|x_t)$), we rearrange terms in \ref{eq:cg} and arrive to following:

\begin{equation}
    \nabla_{x_t}\log p(x_t|y) = \nabla_{x_t}\log p(x_t) +  w\cdot (\nabla_{x_t} \log(x_t|y) - \nabla_{x_t} \log (x_t))
\end{equation}

\paragraph{DCG variants} Now let's derive possible decoupled classifier-free variants for two conditions, specifically when $y = [ c_{text}, c_{id}]$. We note that $\nabla \log p(x_t|c_{text}, c_{id}) - \nabla \log p(x_t)$ from classifier-free guidance corresponds to estimation of $\nabla_{x_t} \log p(c_{id}, c_{text} | x_t)$ score function, which can be expressed in following ways:

\begin{equation}    
\nabla_{x_t} \log p(c_{id}, c_{text} | x_t)  =
\begin{cases}
\nabla_{x_t} \log p(c_{id} | x_t, c_{text}) + \nabla_{x_t} \log p (c_{text} | x_t) \\
\nabla_{x_t} \log p(c_{text} | x_t, c_{id}) + \nabla_{x_t} \log p (c_{id} | x_t) \\
\nabla_{x_t} \log p(c_{id} | x_t) + \nabla_{x_t} \log p (c_{text} | x_t)
\end{cases}
\end{equation}

Last expression is possible if we assume that $p(c_{id}, c_{text}) = p(c_{id})p(c_{text})$, which generally is not true, but since in practice choice of prompts and identities for id-preserving generation are not dependent, it can be valid. Finally, reformulating back to noise prediction, we arrive to three possible DCG formulations, where $DCG_2$ is the one used in main sections of the paper:

\begin{align}
    DCG_1(\hat \epsilon) &:= \epsilon(\varnothing, \varnothing) + \alpha\cdot(\epsilon(c_{text}, \varnothing) - \epsilon(\varnothing, \varnothing) + \beta \cdot (\epsilon(c_{text}, c_{id}) - \epsilon(\epsilon(c_{text}, \varnothing)) \label{eq::dcg-options-1} \\
    DCG_2(\hat \epsilon) &:= \epsilon(\varnothing, \varnothing) + \alpha\cdot(\epsilon(\varnothing, c_{id}) - \epsilon(\varnothing, \varnothing) + \beta \cdot (\epsilon(c_{text}, c_{id}) - \epsilon(\varnothing, c_{id}) \label{eq::dcg-options-2} \\
    DCG_3(\hat \epsilon) &:= \epsilon(\varnothing, \varnothing) + \alpha\cdot(\epsilon(c_{text}, \varnothing) - \epsilon(\varnothing, \varnothing)) + \beta \cdot (\epsilon(\varnothing , c_{id}) - \epsilon(\varnothing, \varnothing)) \label{eq::dcg-options-3}
\end{align}

\begin{figure}[!htb]
  \centering
  \includegraphics[width=0.8\textwidth]{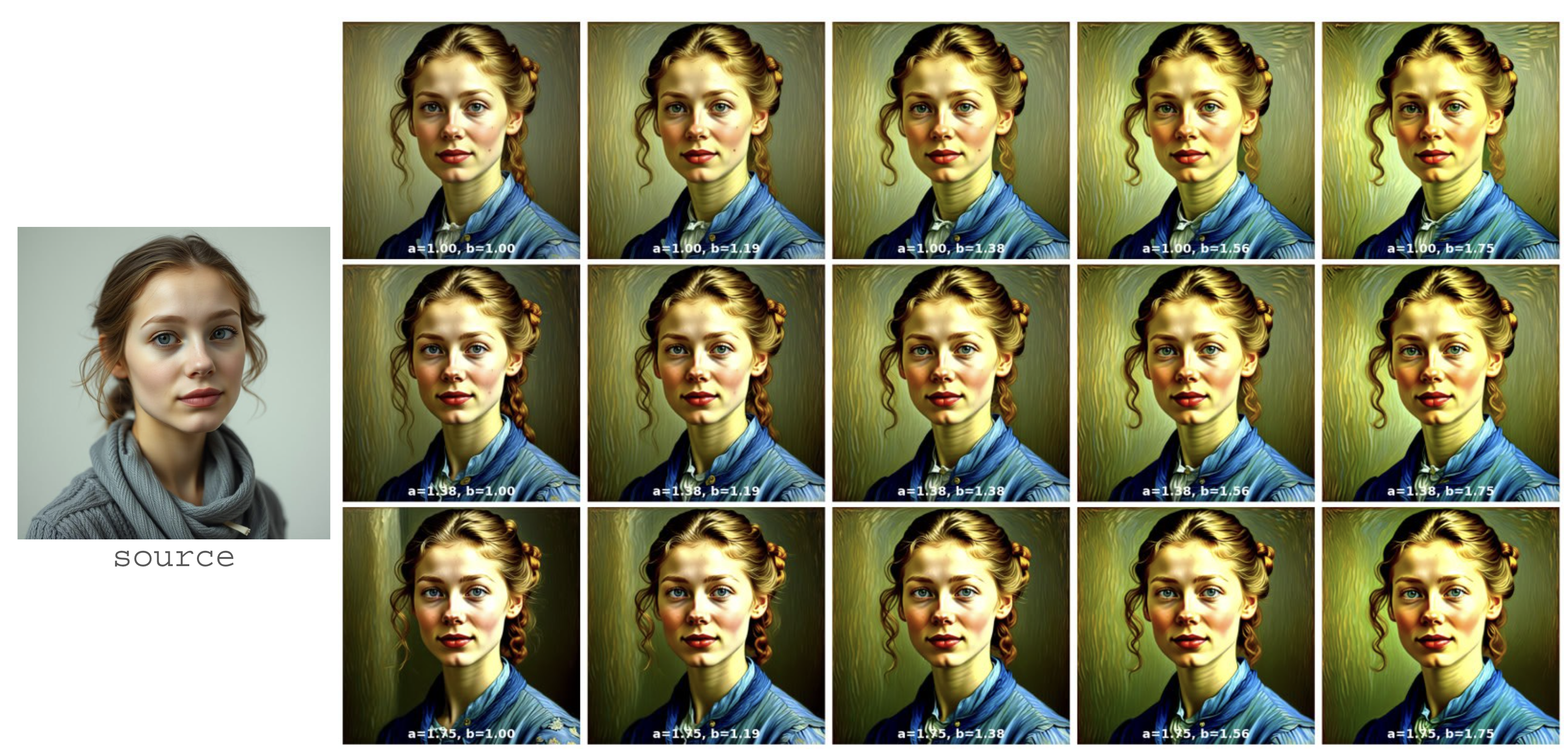}
  \caption{Visual ablation of different coefficient values in DCG ($DCG_2$ from equations above) - larger $a$ values enhance facial features, while $b$ enhance impact of prompt with style. Prompt used for generation was \textit{"van gogh style portrait of a person"}} 
  \label{fig: dcg_coef_grid}
\end{figure}

In practice we find that expression in Eq. \ref{eq::dcg-options-2} works best in terms of semantic changes in the image. In Figure \ref{fig: dcg_coef_grid} we provide ablation of different values for $\alpha$ and $\beta$ and their effect on generation output.

\raggedbottom

\subsection{AM analysis and details}
\label{sec: am-appendix}

\begin{figure}[!htb]
  \centering
  \includegraphics[width=0.8\textwidth]{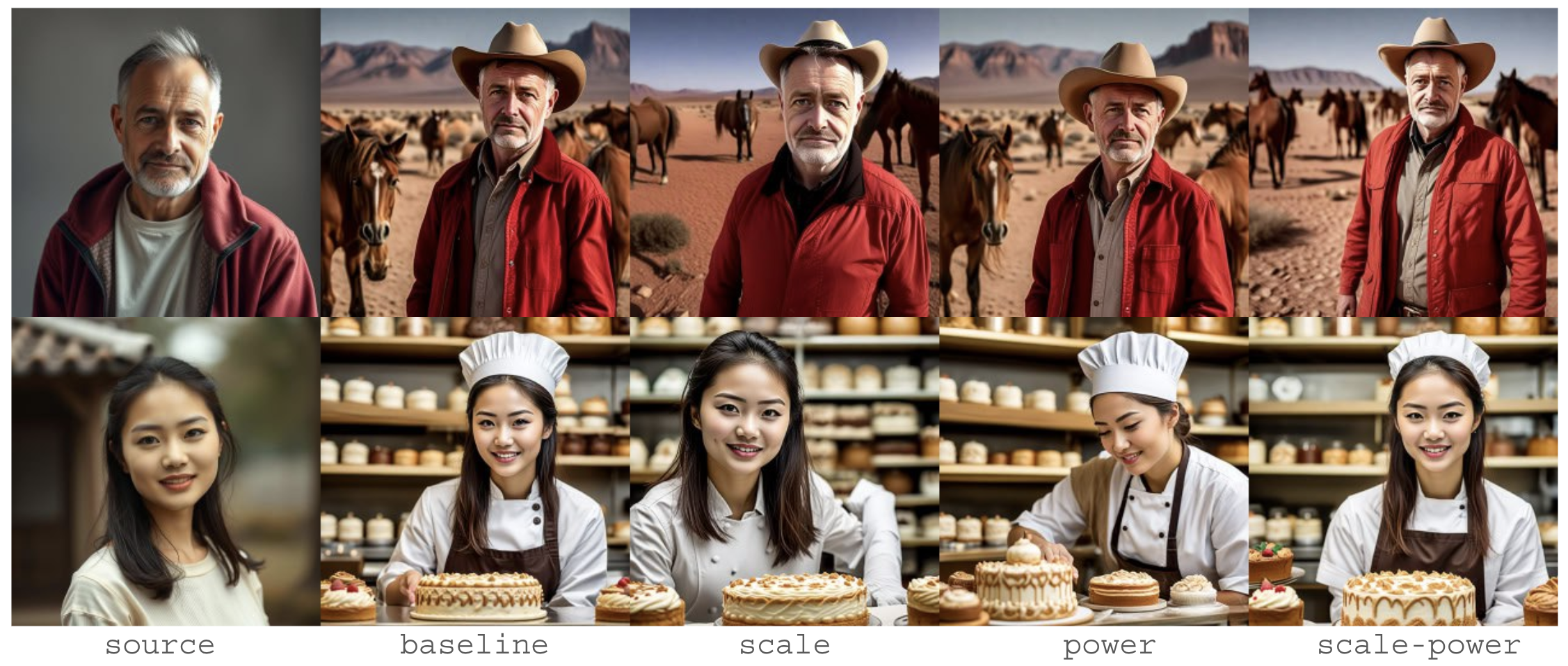}
  \caption{Visual ablation of scale-power transform components} 
  \label{fig: scale-pow-ablate}
\end{figure}

\paragraph{Scale-power ablation} We provide visual ablation why scale-power transformation works in Figure \ref{fig: scale-pow-ablate}. Scaling increases similarity, but alters image background, resulting in prompt following degradation. This is expected, as plugging scaling transform into Eq. \ref{eq::am_definition} instead of $f()$ we can see that it is same as increasing $\lambda$. When raising attention values to some power, we achieve attention values shifting to 0, which decreases identity preservation, but increases prompt following, especially around face, since attention values in decoupled blocks stop interfering with attention from cross-attention blocks. Combination of transforms results in power transform basically canceling prompt following degradation of scale transform.

\raggedbottom

\begin{figure}[!htb]
  \centering
  \includegraphics[width=0.65\textwidth]{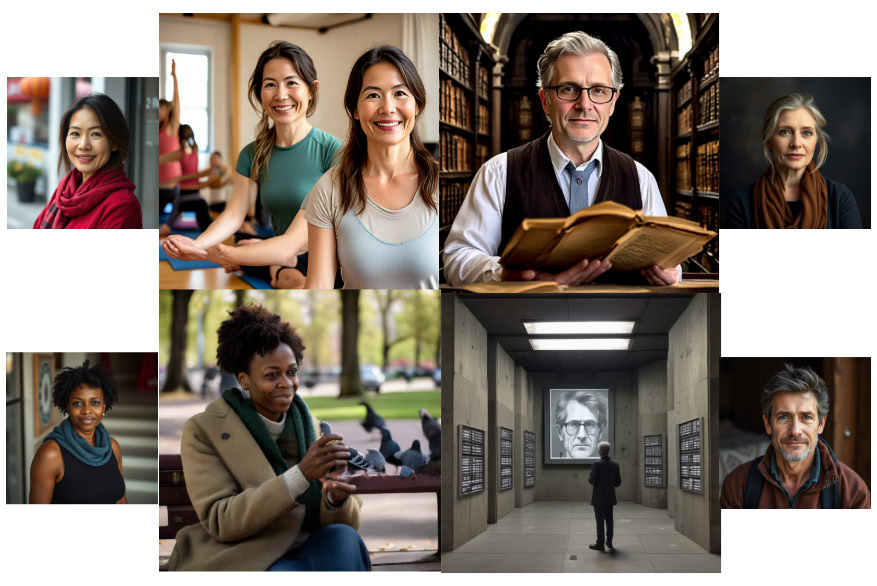}
  \caption{Generation examples with distilled model where generated image fails to successfully preserve identity in meaningful way} 
  \label{fig: failure-cases}
\end{figure}

\paragraph{Failure cases demonstration} In Figure \ref{fig: failure-cases} we give examples of id-preserving failures with distilled diffusion model, where instead of expected outcome with human-centric generation method fails to preserve meaningfully align identity and surrounding context, which can result in identity morphing into background, being  between multiple humans in image, unrealistic postures and etc. Such cases often can't be fixed by proposed scale-power transform, which serves as motivation for a more control-nature transform that changes structure of images.

\raggedbottom

\paragraph{Scheduled-softmask transform details} Beyond details provided in main sections, we also found that attention values for the first token in decoupled CrossAttention (see Fig.\ref{fig: dca-viz}) are inverted - attention is focused on background across all blocks and timesteps, and it's values histogram has mode closer to 1 value. Therefore, when applying transformation to first token, we first invert it's values, and after transform invert back so that AM transformation has same expected effect across all tokens. Secondly, to make \texttt{AdaIN} alignment more stable, we also apply it to transformed decoupled block output with same interpolation hyper-parameter $w$ as one used in Eq. \ref{eq:am_softmask}. 

\begin{figure}[!htb]
    \centering
    \includegraphics[width=0.85\textwidth]{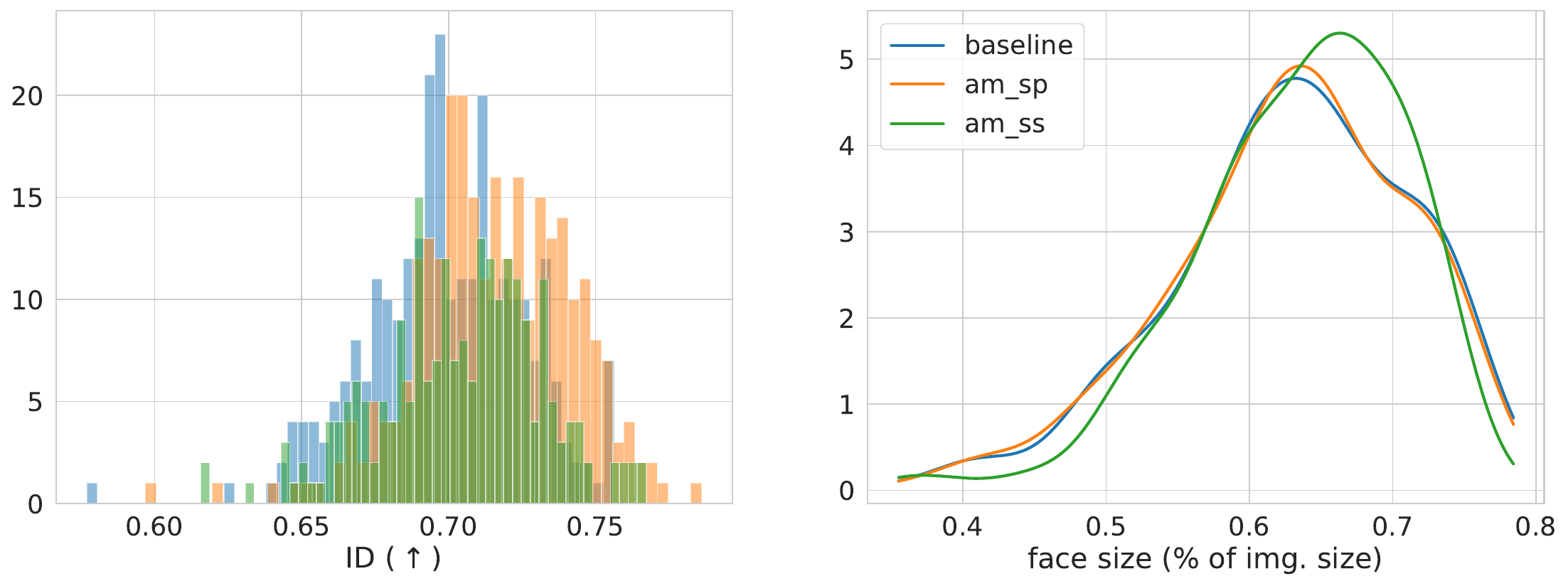}
    \caption{Analysis of transformation application effect on distributions of face similarity on the left and face sizes on the right}
    \label{fig:am_application_stability}
\end{figure}

In Figure \ref{fig:am_application_stability} we evaluate distributions of face similarities to source $c_{id}$ on the left and distribution of face sizes on the right, for number of random generations with fixed arbitrary identity and neutral prompt. This experiments highlight qualitative difference between two AM transforms - both increase identity preservation, while scheduled-softmask being more stable with less variance in ID preservation and generating more large faces without additional control.

\raggedbottom

\subsection{Results of DCG in stylistic setup}
\label{sec: dcg-pareto-fronts-and-tables-appendix}

\begin{figure}[!htb]
  \centering  
    \subfloat[\centering Hyper]{\includegraphics[width=0.5\textwidth]{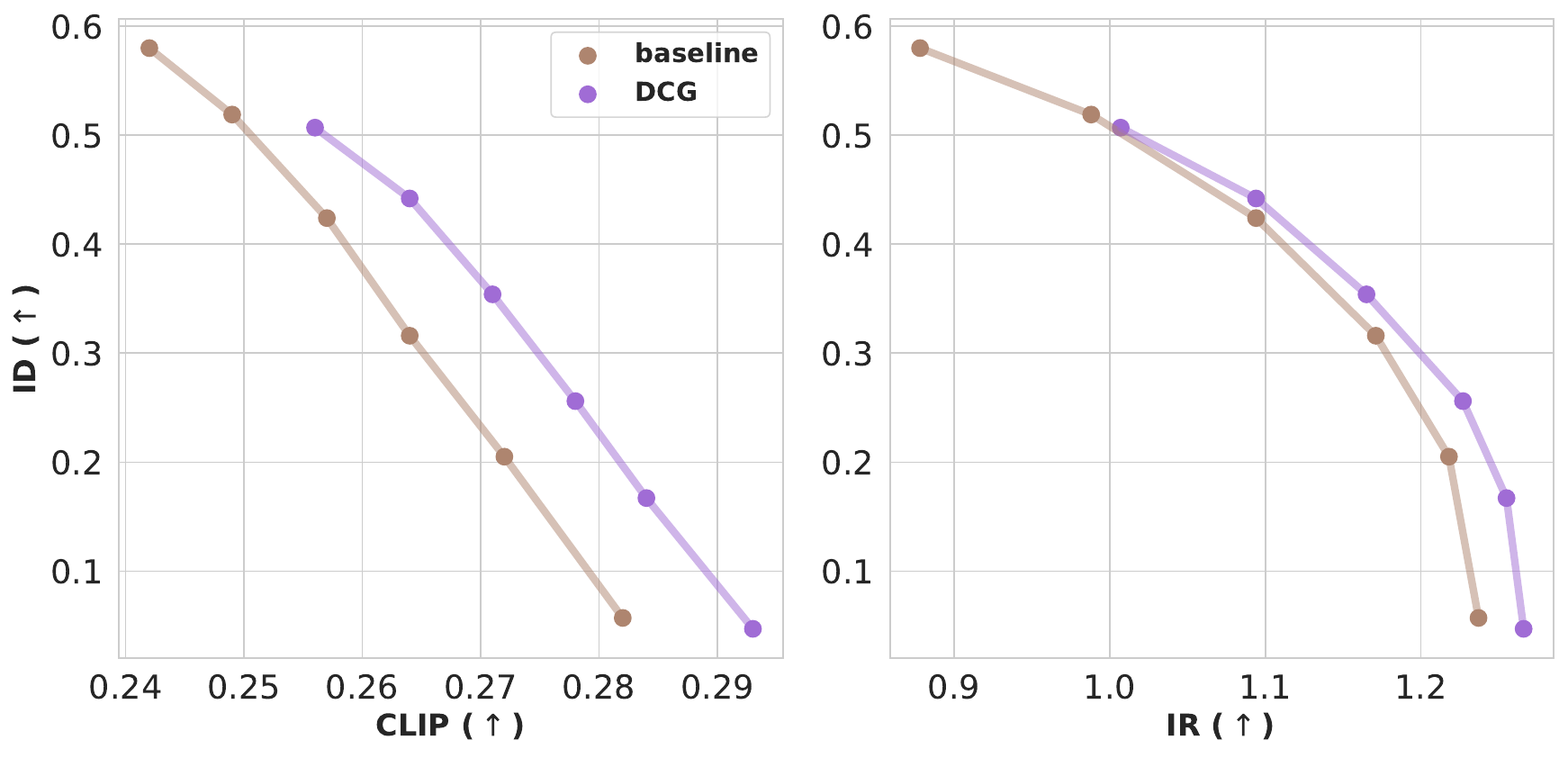}\label{fig:lightning-front-lora1}}
    \subfloat[\centering Lightning]{\includegraphics[width=0.5\textwidth]{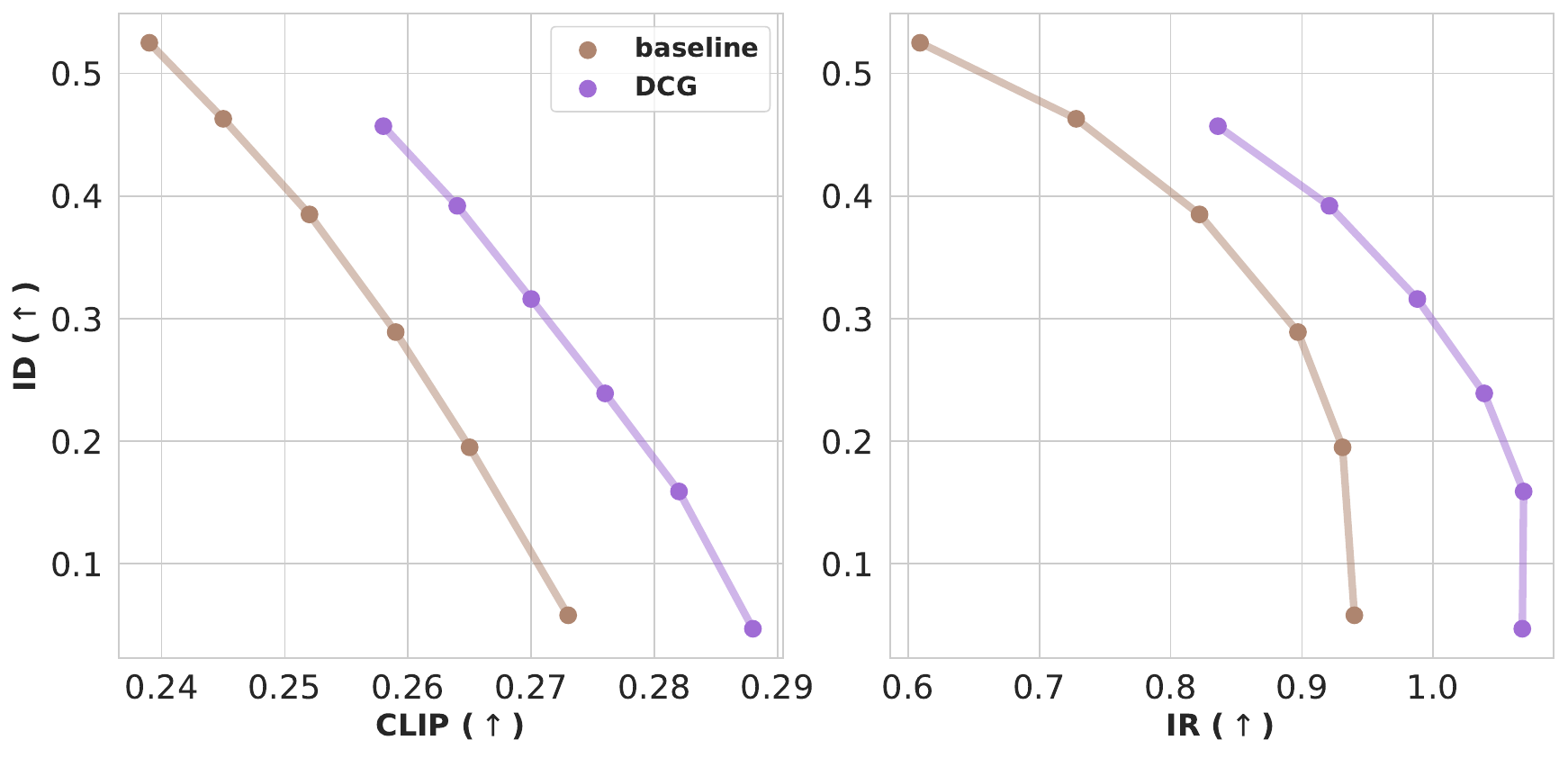}\label{fig:lightning-front-lora05}}
  \caption{Pareto fronts of Hyper and Lightning with DCG against baseline, stylistic setup, \texttt{lora\_scale}$=1$}
  \label{fig::style-fronts}
\end{figure}

In Figure \ref{fig::style-fronts} we present fronts for DCG in stylistic dataset for Hyper and Lightning. Parameters are specified in main section of the text are shares across all models and also joint application with AM.

\raggedbottom

In Table \ref{table: dcg-metrics} we report metric comparisons for fixed \texttt{ip\_adapter\_scale}$=0.8$ for all models. We can observe that DCG achieves expected degradation of face similarity, while increasing CLIP, IR and FCS. 

\begin{table}[!htb]
  \caption{Baseline setups against DCG mechanism}
  \label{table: dcg-metrics}
  \centering
  \begin{tabular}{lllll}
    \toprule
    Model     & \texttt{face\_sim} $\uparrow$ & CLIP-score $\uparrow$  & IR $\uparrow$ & FCS $\uparrow$ \\
    \midrule
    Hyper (base) & \textbf{0.519}  & 0.249 & 0.988 & 0.180   \\
    Hyper + DCG   & 0.442 (-14.8\%) & \textbf{0.264 }(+6.0\%) & \textbf{1.094} (+10.7\%) & \textbf{0.184 }(+2.2\%) \\
    \midrule 
    Lightning (base) & 0.\textbf{463}  & 0.245 & 0.728 & 0.175 \\
    Lightning + DCG  & 0.392 (-15.3\%)  & \textbf{0.264} (+7.8\%) & \textbf{0.921} (+26.5\%) & \textbf{0.181} (+3.4\%) \\
    \midrule
    LCM (base) & \textbf{0.439} & 0.259 & 0.540 & 0.180 \\
    LCM + DCG  & 0.336 (-23.5\%) & \textbf{0.270} (+4.2\%) & \textbf{0.639} (18.3\%) & \textbf{0.181} (+0.6\%) \\
    \midrule
    Turbo (base) & \textbf{0.310} & 0.252 & 0.888 & 0.165 \\
    Turbo + DCG  & 0.254 (-18.1\%) & \textbf{0.277} (+9.9\%) & \textbf{1.007} (+13.4\%) & \textbf{0.175} (+6.1\%) \\
    \bottomrule
  \end{tabular}
\end{table}

\subsection{Results of AMs in realistic setup}
\label{sec: am-pareto-fronts-and-tables-appendix}

Below we present results in terms of fronts computed on realistic subset and full table computed for fixed \texttt{ip\_adapter\_scale}$=0.8$. \texttt{AM1} denotes scale-power transform and \texttt{AM2} denotes scheduled-softmask transform. In all setups (including joint application with DCG in following sections) all hyper-parameters are fixed across checkpoints and are following:

\noindent\underline{AM1} - target "up" and "down" unet parts, power strength $p=1.3$, scale strength $s=1.45$ in "down" part and $s=1.55$ in "up" part.

\noindent\underline{AM2} - target "up" and "down" unet parts, scale strength $s=1.55$ everywhere except first step; softmask quantile $p=0.65$ softmask $d=7.5$ at first step, $d=5.$ at other steps; AdaIN blend coefficient $w=0.7$.

\begin{figure}[!htb]
  \centering  
    \subfloat[\centering Hyper]{\includegraphics[width=0.5\textwidth]{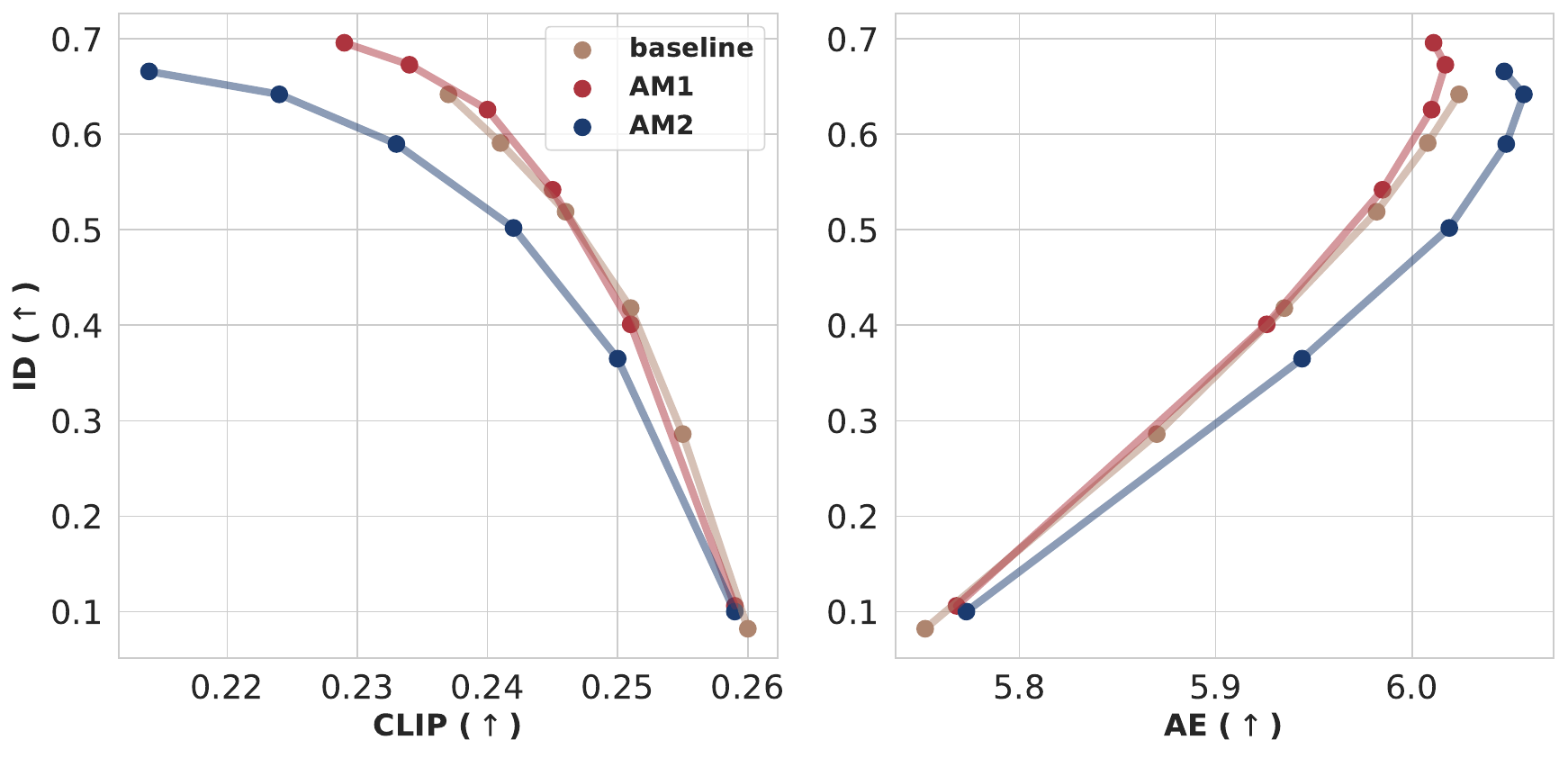}}
    \subfloat[\centering Lightning]{\includegraphics[width=0.5\textwidth]{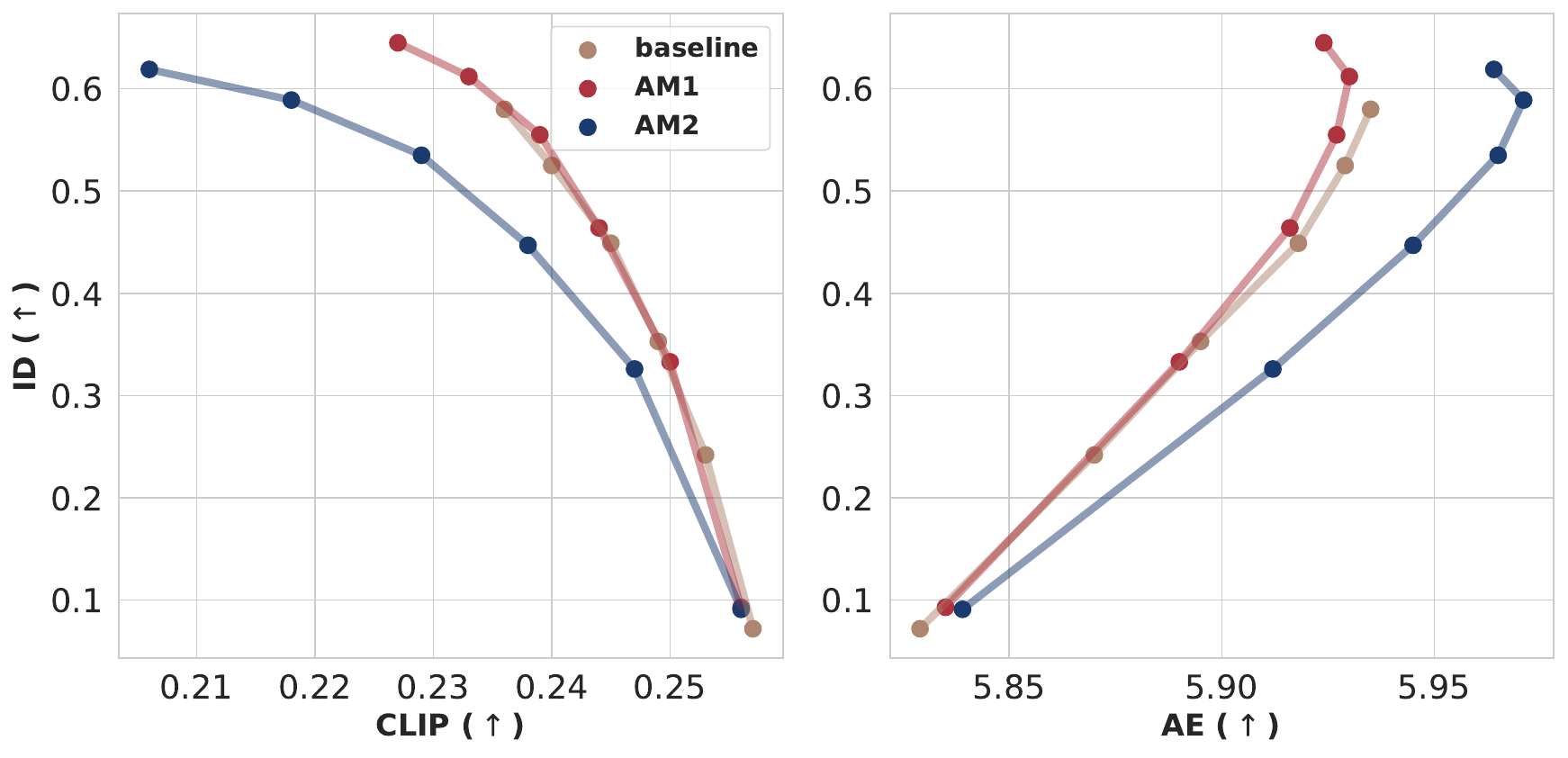}}
  \caption{Pareto fronts of Hyper and Lightning with AM mechanisms against baseline, realistic setup}
  \label{fig::real-fronts}
\end{figure}
\raggedbottom

\begin{table}[!htb]
  \caption{Metric comparison of AM transforms against baseline setups}
  \label{table: am_mechanisms-metrics}
  \centering
  \begin{tabular}{@{}l cccc cccc @{}}
    \toprule
    \multirow{2}{*}{\textbf{Model}} &
    \multicolumn{4}{c}{\texttt{lora\_scale}$=1.0$} &
    \multicolumn{4}{c}{\texttt{lora\_scale}$=0.5$}\\
    \cmidrule(lr){2-5}\cmidrule(lr){6-9}
    & \texttt{ID} $\uparrow$ & CLIP $\uparrow$  & AE $\uparrow$ & FFC $\downarrow$ & \texttt{ID} $\uparrow$ & CLIP $\uparrow$  & AE $\uparrow$ & FFC $\downarrow$ \\
    \midrule
    Hyper (base) & 0.591  & \textbf{0.241} & 6.008 & \textbf{0} & 0.408 & \textbf{0.255} & 6.229 & 19   \\
    Hyper +  $AM1$  & \textbf{0.673} & \underline{0.234} & \underline{6.017} & 1  & \textbf{0.523} & \underline{0.247} & 6.220 & 10 \\
    Hyper +  $AM2$  & \underline{0.642} & 0.224 & \textbf{6.057} & \textbf{0}  & \underline{0.517} & 0.239  &  \textbf{6.265} & \textbf{3} \\
    \midrule 
    Lightning (base) & 0.525  & \textbf{0.240} & 5.929 & 0 & 0.386  & \textbf{0.249} &  6.079 & 18  \\
    Lightning +  $AM1$  & \textbf{0.612}  & \underline{0.233} & \underline{5.930} & 0 & \underline{0.494}  & \underline{0.241} & 6.088 & 12  \\
    Lightning +  $AM2$  & \underline{0.589}  & 0.218 & \textbf{5.971} & 0 & \textbf{0.496}  & 0.231 & \textbf{6.107} & \textbf{1} \\
    \midrule
    LCM (base) & 0.552 & \textbf{0.235} & 5.754 & 3 & 0.380 & \textbf{0.249} & 5.927 & 46 \\
    LCM +  $AM1$ & \textbf{0.610}  & \underline{0.227} & \underline{5.783} & \textbf{1} & \textbf{0.477} & \underline{0.240} & 5.942 & 34 \\
    LCM +  $AM2$ & \underline{0.597}  & 0.214 & \textbf{5.802} & \textbf{1} & \underline{0.476} & 0.231 & \textbf{5.974} & \textbf{18} \\
    \midrule
    Turbo (base) & 0.349 & \textbf{0.243} & \textbf{5.650} & 94 & 0.189 & \textbf{0.250} & 5.764 & 116 \\
    Turbo +  $AM1$  & \textbf{0.467} & \underline{0.235} & 5.635 & 57 & \textbf{0.289} & \underline{0.244} & 5.769 & 63  \\
    Turbo +  $AM2$  & \underline{0.443} & 0.230 & \underline{5.647} & \textbf{62} & \underline{0.283} & 0.240 & \textbf{5.784 } & \textbf{51} \\
    \bottomrule
  \end{tabular}
\end{table}
\raggedbottom


\subsection{Pareto fronts of Lightning model}
\label{sec: full-pareto-fronts-lightning}

Below in Fig. \ref{fig::lightning-full-fronts} we provide additional fronts computed for Lightning model with full FastFace framework on joint evaluation set for varying \texttt{ip\_adapter\_scale} $\in \{0.1, 0.35, 0.5, 0.65, 0.8, 0.95\}$, same values used in main section.

\begin{figure}[!htb]
  \centering  
    \subfloat[\centering \texttt{lora\_scale}$=1.0$]{\includegraphics[width=0.5\textwidth]{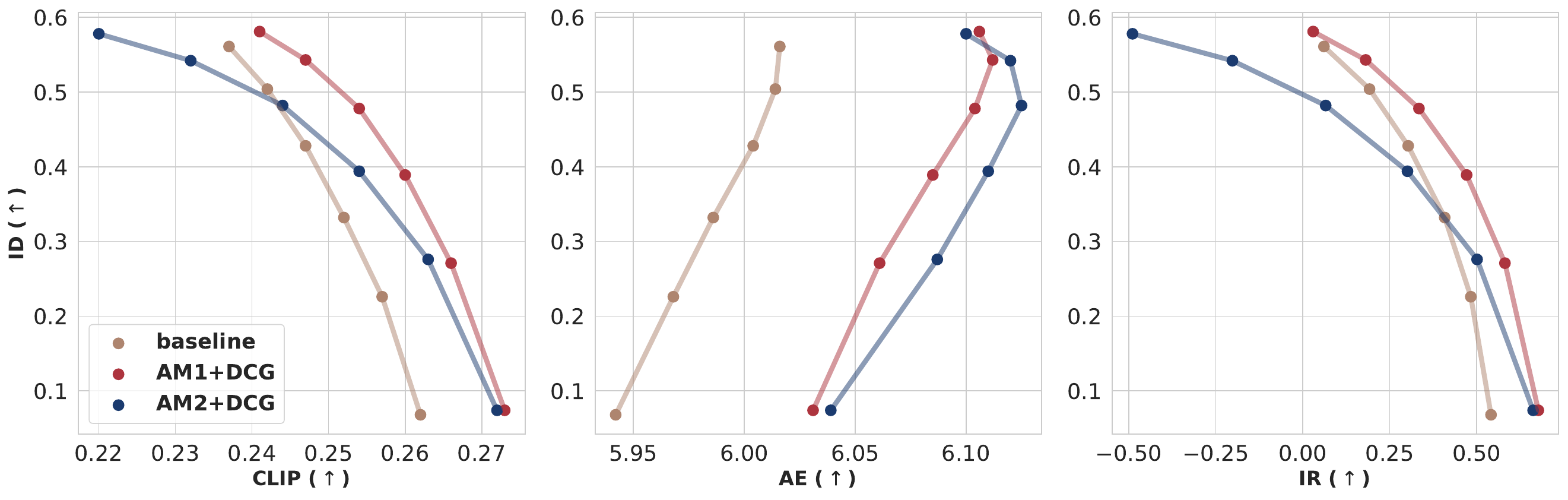}}
    \subfloat[\centering \texttt{lora\_scale}$=0.5$]{\includegraphics[width=0.5\textwidth]{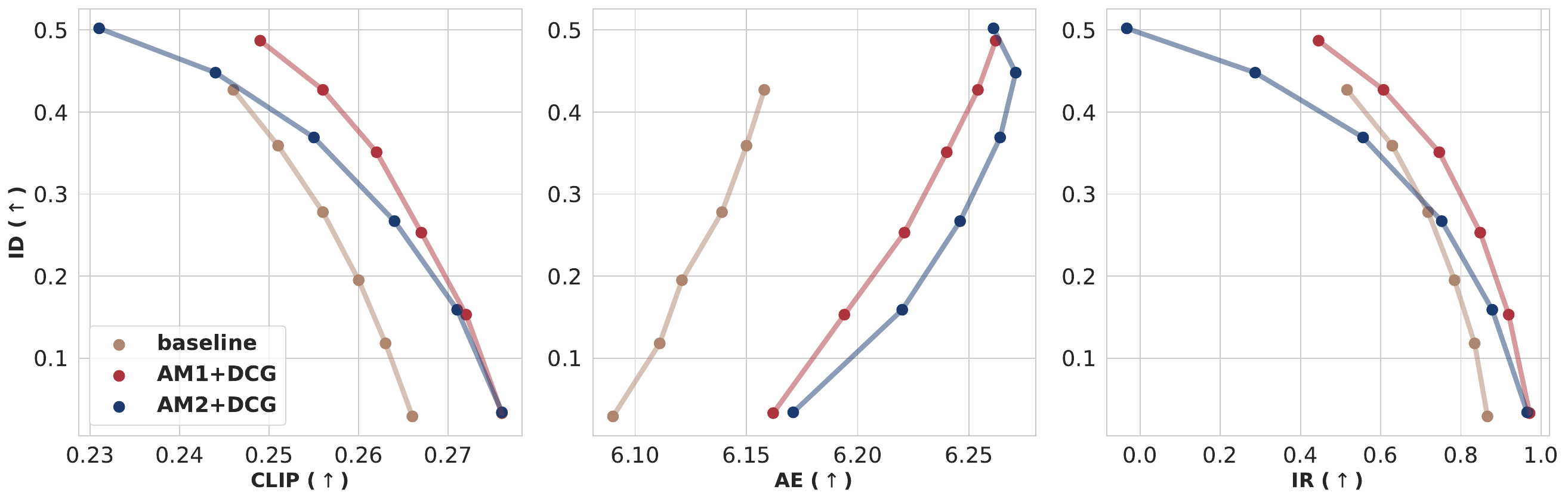}}
  \caption{Lightning fronts for full data setup, different FastFace configurations and \texttt{lora\_scales}}
  \label{fig::lightning-full-fronts}
\end{figure}
\raggedbottom

\subsection{Qualitative results with different models}
\label{sec: methods-visual-results-appendix}

In figures below we visualize additional demonstrations of AM mechanisms for LCM and Turbo checkpoints, as well as general comparison for different prompt categories with a lot of different setups to highlight differences that proposed methods introduce. For notational convenience with main sections of paper here we denote scale-power transform as \texttt{AM1} and scheduled-softmask as \texttt{AM2}, as well as joint applications with DCG as $FF$ with according subscript.

\begin{figure}[!htb]
  \centering
  \includegraphics[width=0.85\textwidth]{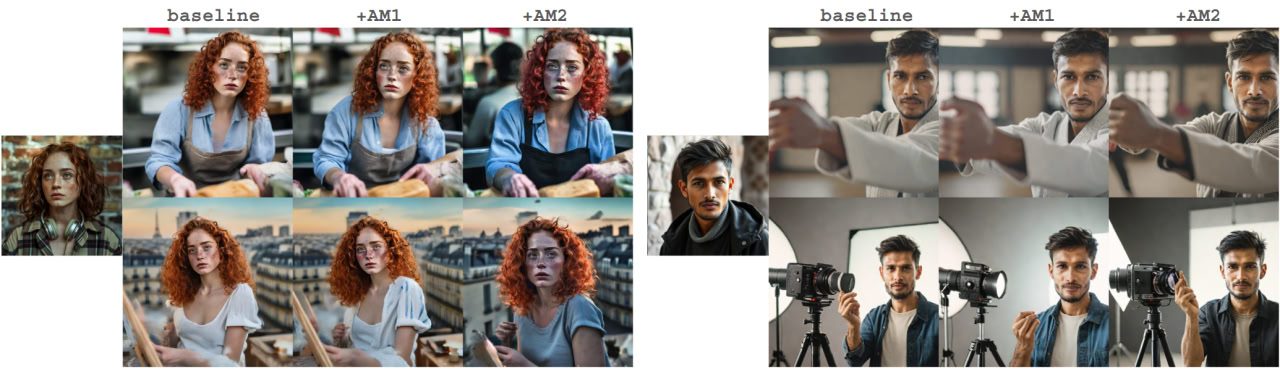}
  \caption{AM with LCM and Turbo checkpoints, \texttt{lora\_scale}$=1.$} 
\label{fig: lcmturbo_fulllora}
\end{figure}
\raggedbottom

\begin{figure}[!htb]
  \centering
  \includegraphics[width=0.85\textwidth]{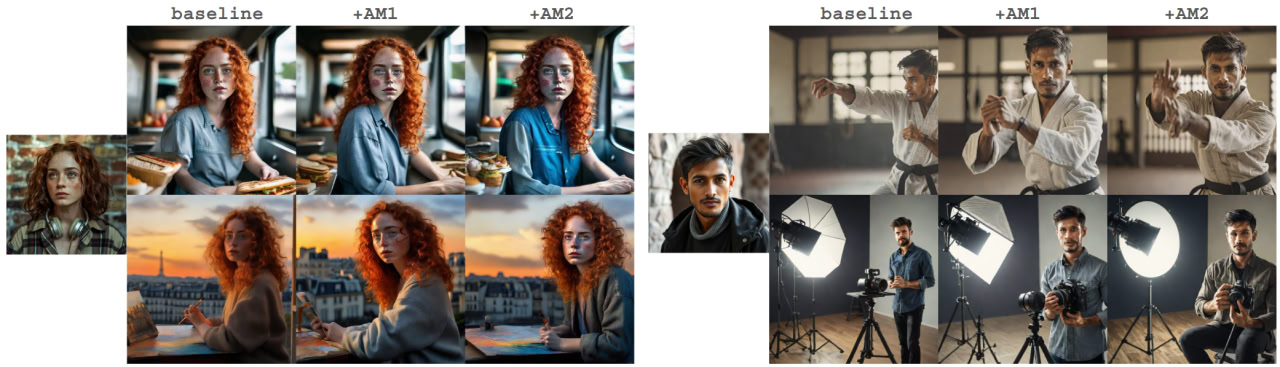}
  \caption{AM with LCM and Turbo checkpoints, \texttt{lora\_scale}$=0.5$} 
\label{fig: lcmturbo_lowlora}
\end{figure}
\raggedbottom

\begin{figure}[!htb]
  \centering
  \includegraphics[width=0.85\textwidth]{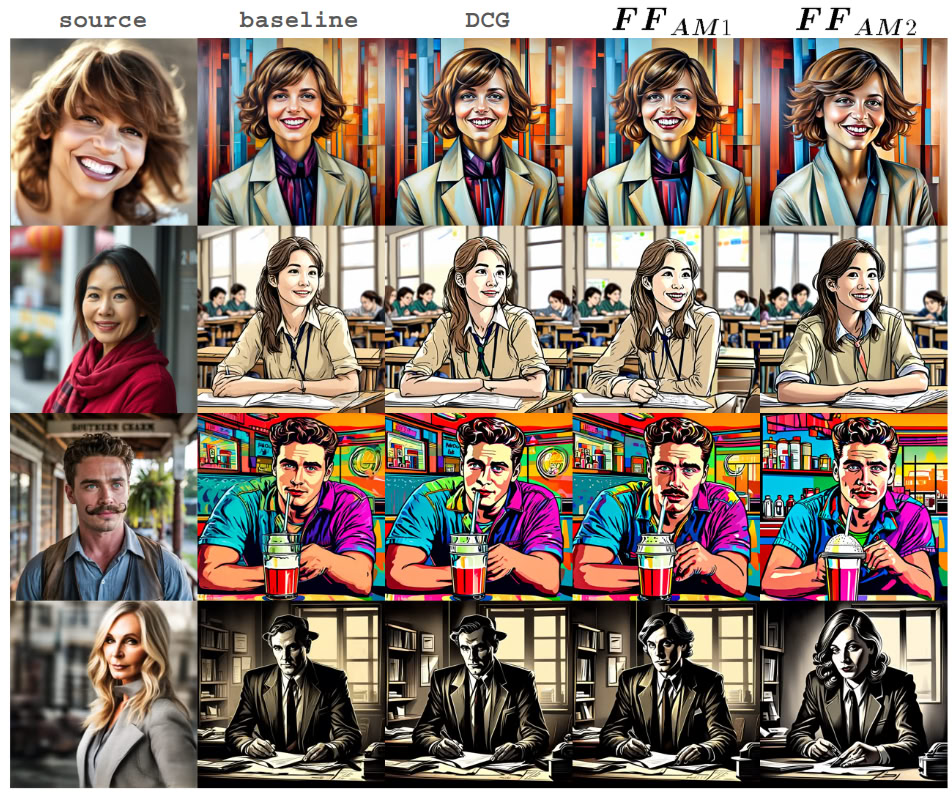}
  \caption{Demonstration of different configurations with stylistic prompts, Hyper checkpoint} 
\label{fig: hyper_configs_style_quality}
\end{figure}
\raggedbottom

\begin{figure}[!htb]
  \centering
  \includegraphics[width=0.85\textwidth]{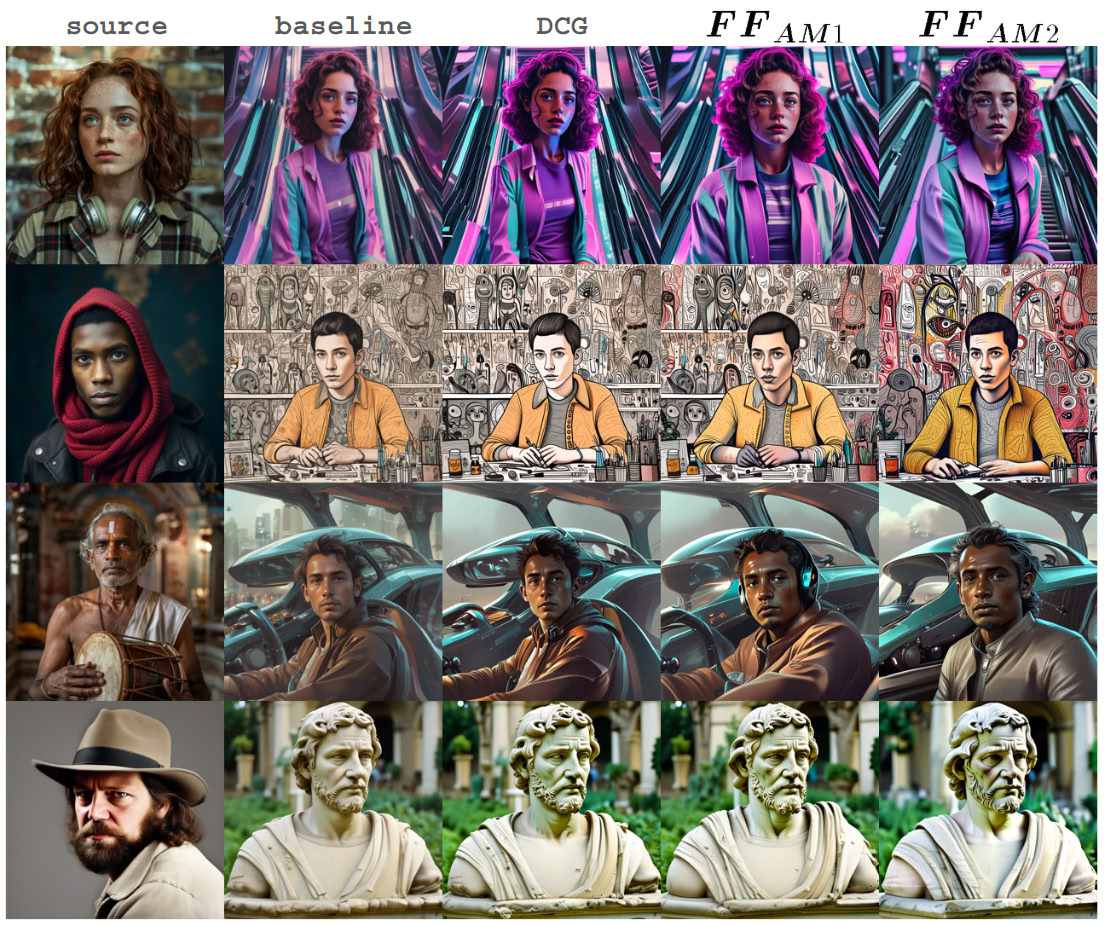}
  \caption{Demonstration of different configurations with stylistic prompts, Lightning checkpoint} 
\label{fig: lightning_configs_style_quality}
\end{figure}
\raggedbottom

\begin{figure}[!htb]
  \centering
  \includegraphics[width=0.85\textwidth]{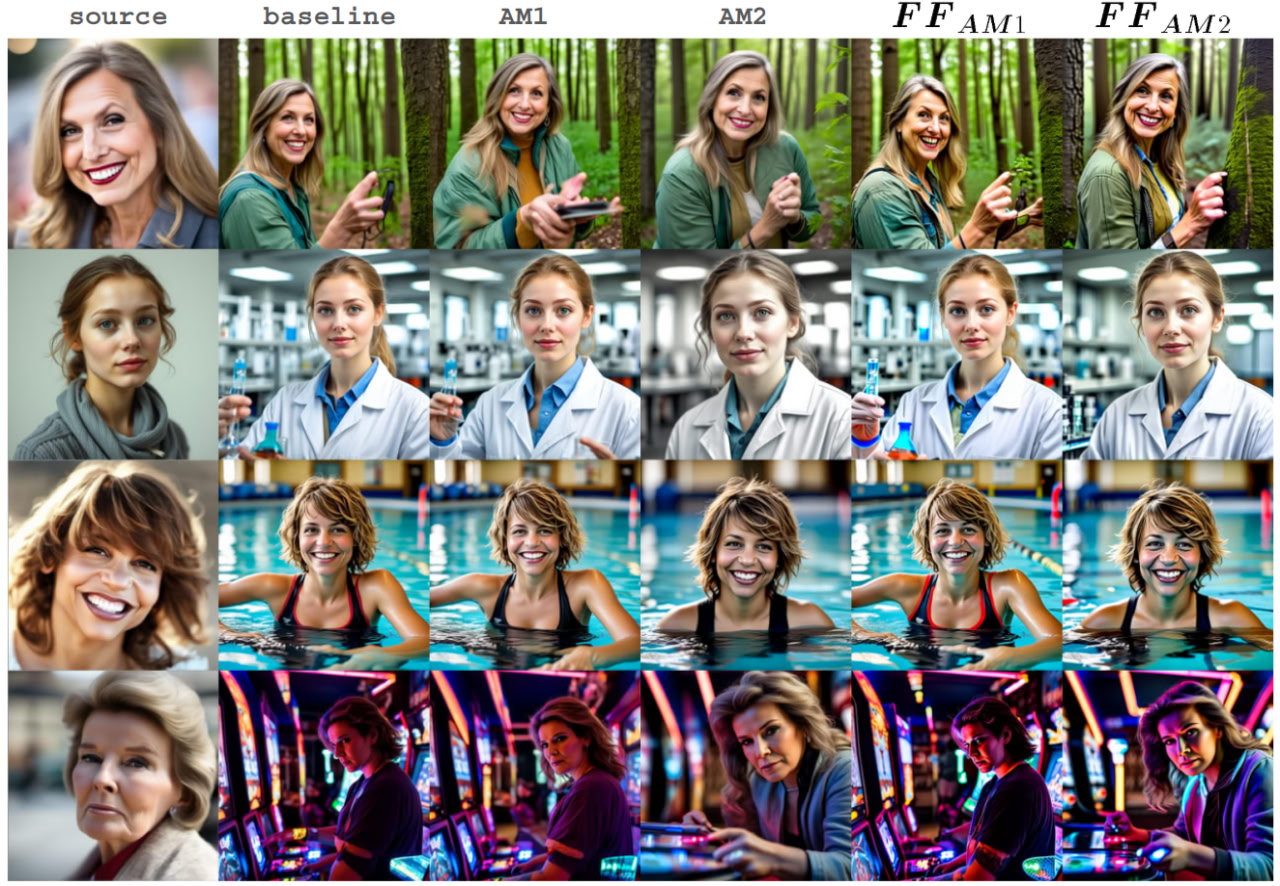}
  \caption{Demonstration of different configurations with realistic prompts, Hyper checkpoint} 
\label{fig: hyper_configs_real_quality}
\end{figure}
\raggedbottom

\begin{figure}[!htb]
  \centering
  \includegraphics[width=0.85\textwidth]{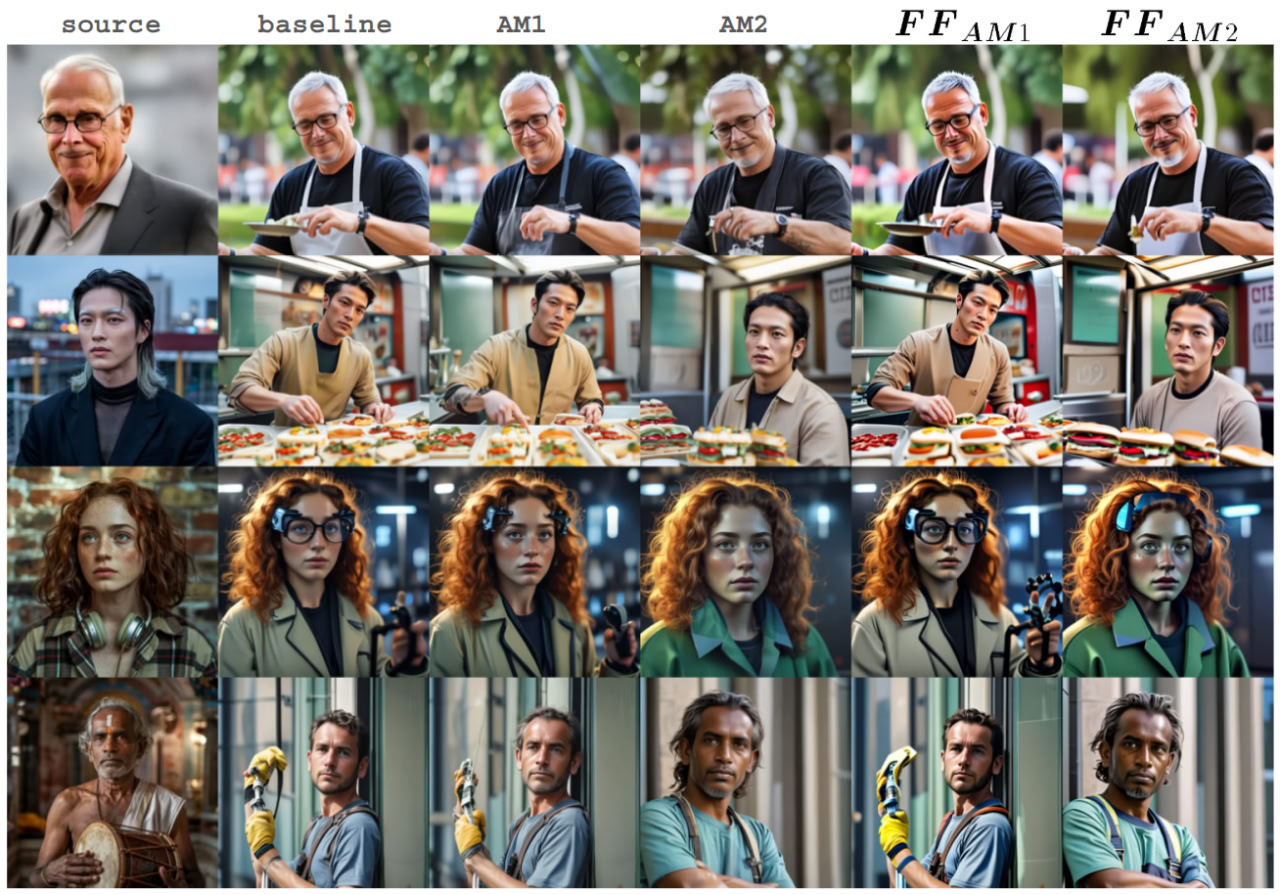}
  \caption{Demonstration of different configurations with realistic prompts, Lightning checkpoint} 
\label{fig: lightning_configs_real_quality}
\end{figure}
\raggedbottom

\begin{figure}[!htb]
  \centering
  \includegraphics[width=0.85\textwidth]{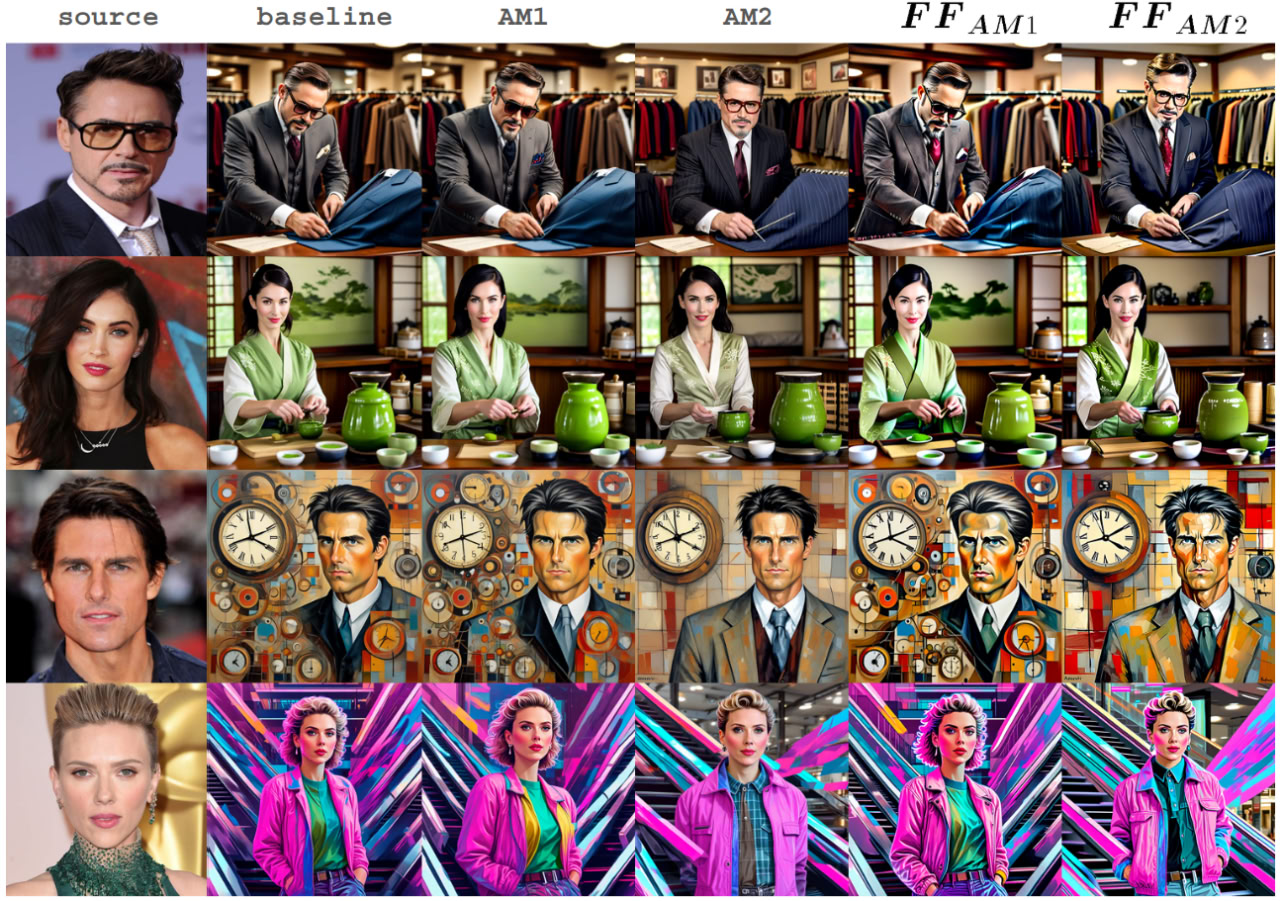}
  \caption{Demonstration of application to images of real people (presented identities are not part of evaluation dataset), Hyper checkpoint} 
\label{fig: realids_configs_quality}
\end{figure}
\raggedbottom

\end{document}